\documentclass[12pt]{mytmp}

\usepackage{fullpage}
\usepackage[utf8]{inputenc} 
\usepackage[T1]{fontenc}    
\usepackage{hyperref}       
\usepackage{url}            
\usepackage{booktabs}       
\usepackage{amsfonts}       
\usepackage{nicefrac}       
\usepackage{microtype}      
\usepackage{xcolor}
\usepackage{graphicx}
\usepackage{subfigure} 
\usepackage{multirow}
\usepackage{amsmath}
\usepackage{amssymb, bm}
\usepackage{tablefootnote}
\usepackage{lineno}
\usepackage{pdflscape}
\usepackage{tabularx}
\usepackage{caption}

\begin{document} 

\begin{frontmatter}

\title{Lifelike Agility and Play in Quadrupedal Robots using Reinforcement Learning and Generative Pre-trained Models}

\author{Lei Han$^{*\dag}$\footnotetext{$^{*}$\text{Equal contribution.}}\footnotetext{$^{\dag}$\text{Correspondence: \url{leihan.cs@gmail.com}, \url{qingxuzhu@tencent.com}.}}}
\author{Qingxu Zhu$^{*\dag}$}
\author{Jiapeng Sheng$^{*}$}
\author{Chong Zhang$^{*}$}
\author{Tingguang Li$^{*}$}
\author{Yizheng Zhang$^{*}$}
\author{He Zhang$^{*}$}
\author{Yuzhen Liu}
\author{Cheng Zhou}
\author{Rui Zhao}
\author{Jie Li}
\author{Yufeng Zhang}
\author{Rui Wang}
\author{Wanchao Chi}
\author{Xiong Li}
\author{Yonghui Zhu}
\author{Lingzhu Xiang}
\author{Xiao Teng}
\author{Zhengyou Zhang}

\address{Tencent Robotics X, Shenzhen, China}
\address{\url{https://tencent-roboticsx.github.io/lifelike-agility-and-play/}}
\address{Published in Nature Machine Intelligence, 2024}


\begin{abstract}
Knowledge from animals and humans inspires robotic innovations. Numerous efforts have been made to achieve agile locomotion in quadrupedal robots through classical controllers or reinforcement learning approaches. These methods usually rely on physical models or handcrafted rewards to accurately describe the specific system, rather than on a generalized understanding like animals do. Here we propose a hierarchical framework to construct primitive-, environmental- and strategic-level knowledge that are all pre-trainable, reusable and enrichable for legged robots. The primitive module summarizes knowledge from animal motion data, where, inspired by large pre-trained models in language and image understanding, we introduce deep generative models to produce motor control signals stimulating legged robots to act like real animals. Then, we shape various traversing capabilities at a higher level to align with the environment by reusing the primitive module. Finally, a strategic module is trained focusing on complex downstream tasks by reusing the knowledge from previous levels. We apply the trained hierarchical controllers to the MAX robot, a quadrupedal robot developed in-house, to mimic animals, traverse complex obstacles and play in a designed challenging multi-agent chase tag game, where lifelike agility and strategy emerge in the robots.
\end{abstract}


\end{frontmatter}


\section{MAIN}

Animals demonstrate swift and graceful movements and precision in response to the environment. Understanding and mimicking animal behaviours can benefit robotic design and control. Legged robots, a common type of bio-inspired robot have been a substantial area of interest in robotic research for decades.

Recently, learning-based approaches have shown promise in legged robot control. Learning-based methods are fully automated and the controllers can be optimized in an end-to-end fashion from robot sensor readings to motor control signals. For example, simulation-based deep reinforcement learning (RL) has been applied in learning legged locomotion over various terrains~\cite{tan2018sim, haarnoja2018latent, hwangbo2019learning,doi:10.1126/scirobotics.abc5986, doi:10.1126/scirobotics.abk2822, kumar2021rma,cheng2023extreme,zhuang2023robot,hoeller2023anymal,yang2023cajun,caluwaerts2023barkour}.
These approaches generally adopt deep RL algorithms to train locomotion tasks in simulation and then apply the trained controllers to legged robots in reality. Contrastingly, efforts have also focused on reducing the gap between simulation and reality, by using hybrid perceptive sensors~\cite{doi:10.1126/scirobotics.abk2822}, accurate modelling of the terrain dynamics~\cite{choi2023learning} and adaptation techniques~\cite{kumar2021rma}.
Moreover, considering more agile locomotion behaviors, ref. \cite{yang2020multi} proposes a mixture-of-expert architecture to produce versatile locomotive skills, and some existing approaches have already tried to mimic animal behaviours from motion capture data. For example, refs. \cite{RoboImitationPeng20,bohez2022imitate} use an imitation learning framework to track motion data collected from real-world animals and then apply the trained controller to real legged robots.
These methods demonstrate that using imitation learning, the trained controller could drive the legged robot to exhibit animal-like movements for general locomotion skills like walking and running. In this study, we aim to push the frontier of legged robot control methods to understand animal behaviours through deep pre-trained representations and generalize lifelike behaviours to solve complex tasks.

Another parallel, but closely related, research field focusing on producing naturalistic behaviours is computer graphics. For example, compressing motion data into reusable priors has been investigated using a latent variable model in character animation~\cite{levine2012continuous, ling2020character}. 
These approaches assume a continuous Gaussian space for the latent embedding learned from motion data and then navigate this space to accomplish user-specified tasks. In ref.~\cite{tirumala2022behavior}, the authors present a perspective to introduce knowledge priors into an RL problem and selectively generalize certain aspects of the learned behaviours. More systematically, a hierarchical architecture has been adopted by stacking multi-level skills sequentially to solve downstream tasks~\cite{haarnoja2018latent,heess2016learning}. 
More recent evidences on physics-based character control show that hierarchical policies could address challenging tasks ranging from complex motor control~\cite{merel2018neural,hasenclever2020comic} to multi-agent cooperation~\cite{liu2022motor,zhu2023ncp}. 
However, all these works are demonstrated in simulators. Another recent study has presented a hierarchical framework to enable soccer shooting by a legged robot~\cite{ji2022hierarchical}, while it focuses on solving a specific task and does not consider the generalization of agile behaviours like animals. Impressed by the power of large deep generative models in understanding images and language, here we explore whether the pre-training idea can be incorporated in robot control. Deep generative models trained on massive datasets can shape expressive representations that can be reused in new tasks. Motivated by this point, instead of straightforwardly training an RL controller towards solving specific locomotion tasks, we initially trained a primitive motor controller to reproduce all animal motions in the dataset using RL, without knowing any exteroceptive information and without focusing on any specific tasks or skills. We defer the understanding of exteroceptive information and the learning of specialist skills to the upper-level training that will be introduced later. The primitive motor controller is similar to previous approaches~\cite{RoboImitationPeng20,bohez2022imitate}, but differing from them, we propose Vector Quantized Primitive Motor Control (VQ-PMC), a novel pre-training method for shaping primitive knowledge from animals in a discrete latent embedding space. 
VQ-PMC brings the spirit in the Vector Quantized Variational AutoEncoders (VQ-VAE, initially devised for generating high-quality images)~\cite{van2017neural,ramesh2021zero} for learning discrete representations from vector-quantized embeddings, and it generalizes our concurrent work in neural categorical priors~\cite{zhu2023ncp} for real legged robots.
In contrast to VQ-VAE, which reconstructs input samples, VQ-PMC generates control signals conditioned on certain robot proprioceptive states. The vector quantized representations serve as a core part of the primitive motor controller, which is trained using RL to reproduce behaviours from captured motions of a Labrador retriever.
As we will demonstrate, the discrete representations from VQ-PMC are highly expressive to compress critical information in the animal motion data, compared with state-of-the-art methods~\cite{van2017neural,ramesh2021zero} that use Gaussian latent representations. In fact, the vector-quantized representations share a similarity with Gaussian mixtures theoretically~\cite{roy2018theory,bishop2006pattern}.
The VQ-PMC is designed to take only proprioception of the robot without knowing the exteroceptive environment. At the successive stage, we reuse the decoder part of the pre-trained primitive motor controller and build upon it with another network that is perceptual of the environment.

The environmental-level network outputs a categorical distribution over a number of latent embeddings trained previously to drive the primitive decoder to output control signals. The difference is that the environmental-level network additionally receives exteroceptive information and a command on movement direction and speed. Then, the combination of the environmental- and primitive-level networks consists of a valid controller, referred to as the environmental–primitive motor controller (EPMC). The EPMC is trained to solve a number of challenging traversing tasks, including creeping through narrow spaces, jumping over multiple hurdles and freerunning over multiple scattered blocks. After this stage of training, knowledge for understanding exteroceptive information, the command constraint and the way to drive the VQ-PMC is preserved in the environmental-level network parameters. Finally, based on the two pre-trained neural networks, we create a strategic-level network to solve a specific task, that is, a challenging multi-agent chase tag game. The strategic-level network receives additional task-specific features and outputs a command on movement direction and speed to drive the lower level controllers. It is worth mentioning that although the strategic-level network locates at the upper most level, it is still reusable in our design. For example, when we want to traverse new obstacles not considered in the current scope of this study, we only need to enrich the capability of the environmental-level network with policy distillation and RL fine-tuning methods, without the need to touch the trained strategic- and primitive-level networks. The multi-level training stages form a general framework that divides the learning into three parts, which focus on proprioceptive, exteroceptive and strategic information, resulting in highly orthogonal multi-level prior knowledge for flexible re-usage.

Following the proposed framework, we obtain a complete hierarchical controller and apply it to the quadrupedal robot MAX~\cite{zhou2023max,chi2022linearization} in a zero-shot manner. We observe that the MAX robot demonstrates animal-level lifelike agility in real-world tasks. MAX can successfully traverse all the considered scenarios in reality. Specifically, we design an interesting yet challenging multi-agent game, in which two MAX robots participate in a chase tag game to demonstrate animal-level strategy. We also conduct comprehensive ablation studies to evaluate the effectiveness of core components to demonstrate the advantage of the proposed framework.

\section{FRAMEWORK OVERVIEW}

\begin{figure}[t]
\centering
\includegraphics[width=\linewidth]{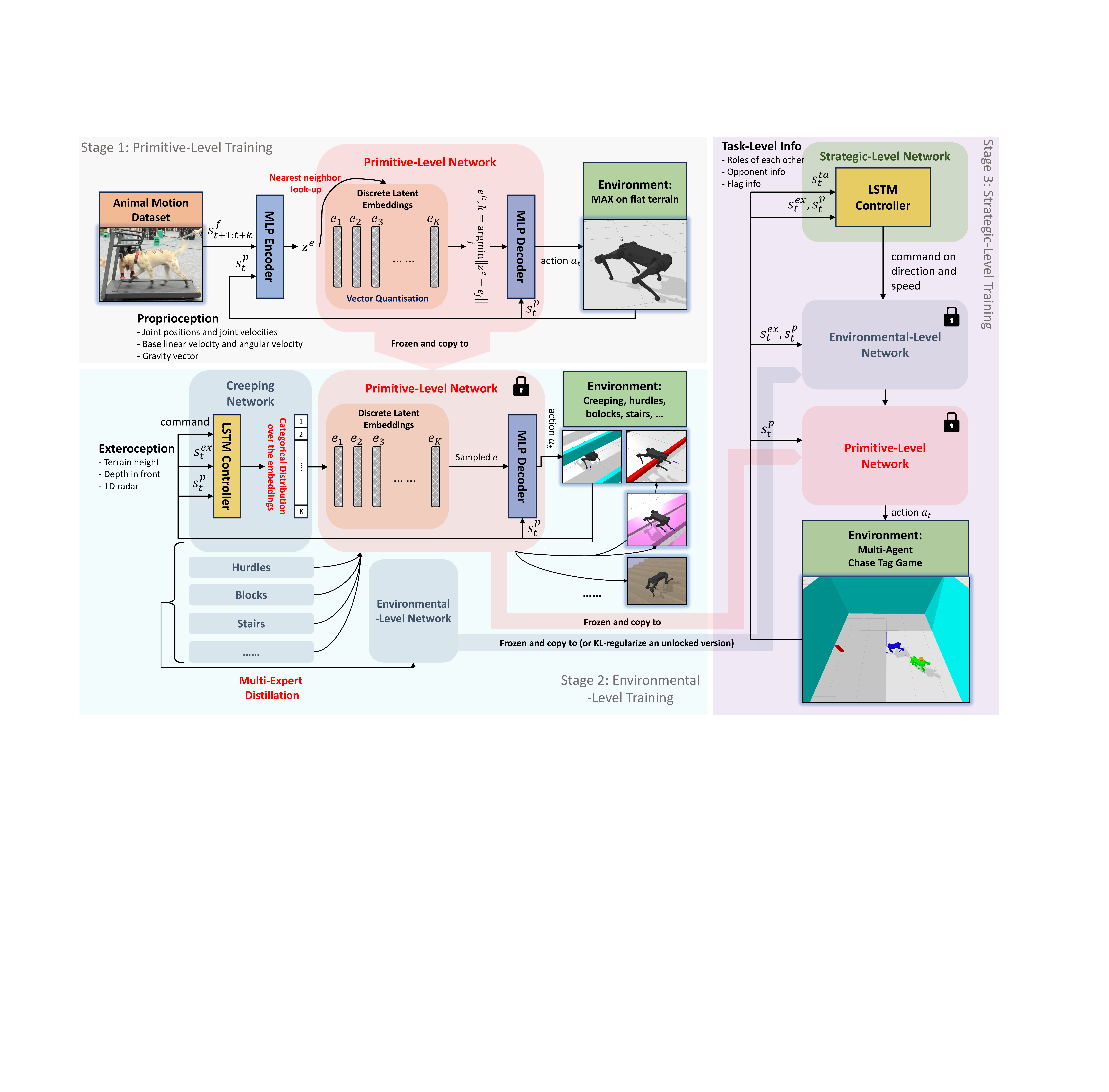}
\caption{\small A framework overview of the proposed method. We initially train a PMC to imitate animal movements using discrete latent embeddings (Stage 1). The decoder of PMC is reused to train environmental-level controllers for general walking, fall recovery, creeping over narrow space, and traversing over hurdles, blocks and stairs separately, which are compressed into a uniform environmental-level controller by multi-expert distillation (Stage 2). At the final stage, we reuse the pre-trained environmental- and primitive-level networks to train a strategic-level network for solving a designed multi-agent chase tag game (Stage 3).
} 
\label{fig:overview}
\end{figure}

An overview of the proposed framework is depicted in Figure~\ref{fig:overview}. The method consists of three stages of reinforcement learning, each of which focuses on extracting knowledge at a different level of task and perception. To avoid confusion, the complete end-to-end controller at each learning stage is referred to as the primitive motor controller (PMC), environmental–primitive motor controller (EPMC) and the strategic–environmental–primitive motor controller (SEPMC), respectively. The partial network that is reusable from each stage of learning is referred to as the primitive-level network, the environmental-level network and the strategic-level network, respectively. As shown in Figure~\ref{fig:overview}, we initially train the PMC to imitate animal movements by using discrete latent embeddings. The trained primitive-level network is frozen and reused to train several environmental-level networks, 
which take additional exteroception and command on direction and speed to output a latent embedding to drive the primitive-level network. 
The sampled embedding will trigger the primitive-level network to generate motor actions to control the robot. We train several environmental-level networks for locomotion on flat terrain, creeping, and traversing over hurdles, blocks and stairs, separately. These environmental-level networks are then compressed into a uniform environmental-level network through multi-expert distillation. 
At the final stage, we reuse the pre-trained primitive- and environmental-level networks to train a complete SEPMC for solving a designed multi-agent chase tag game. The strategic-level network takes additional task-specific information to output commands on direction and speed for the environmental-level network. All the stages are trained using RL in PyBullet simulation, and we adopt the PPO~\cite{schulman2017proximal} algorithm under the distributed multi-agent RL architecture TLeague~\cite{sun2020tleague} throughout the paper. All trained controllers are deployed in real robots in a zero-shot manner.

\section{RESULTS}
Supplementary Videos~1-3 summarize the overall results of the proposed framework. We deployed the trained controllers on the MAX robot~\cite{zhou2023max,chi2022linearization}, which is a four-legged robot weighted 14 kg. Each leg of MAX contains three actuators that can provide continuous torque of 22 Nm on average and 30 Nm maximum. All the training tasks are completed with up to four NVIDIA TESLA V100 GPUs and 4 days of training, including imitation, traversing, distillation and self-play (for the chase tag game) tasks.

\subsection{Primitive Behaviors}

\begin{figure}[h]
\centering
\includegraphics[width=0.99\linewidth]{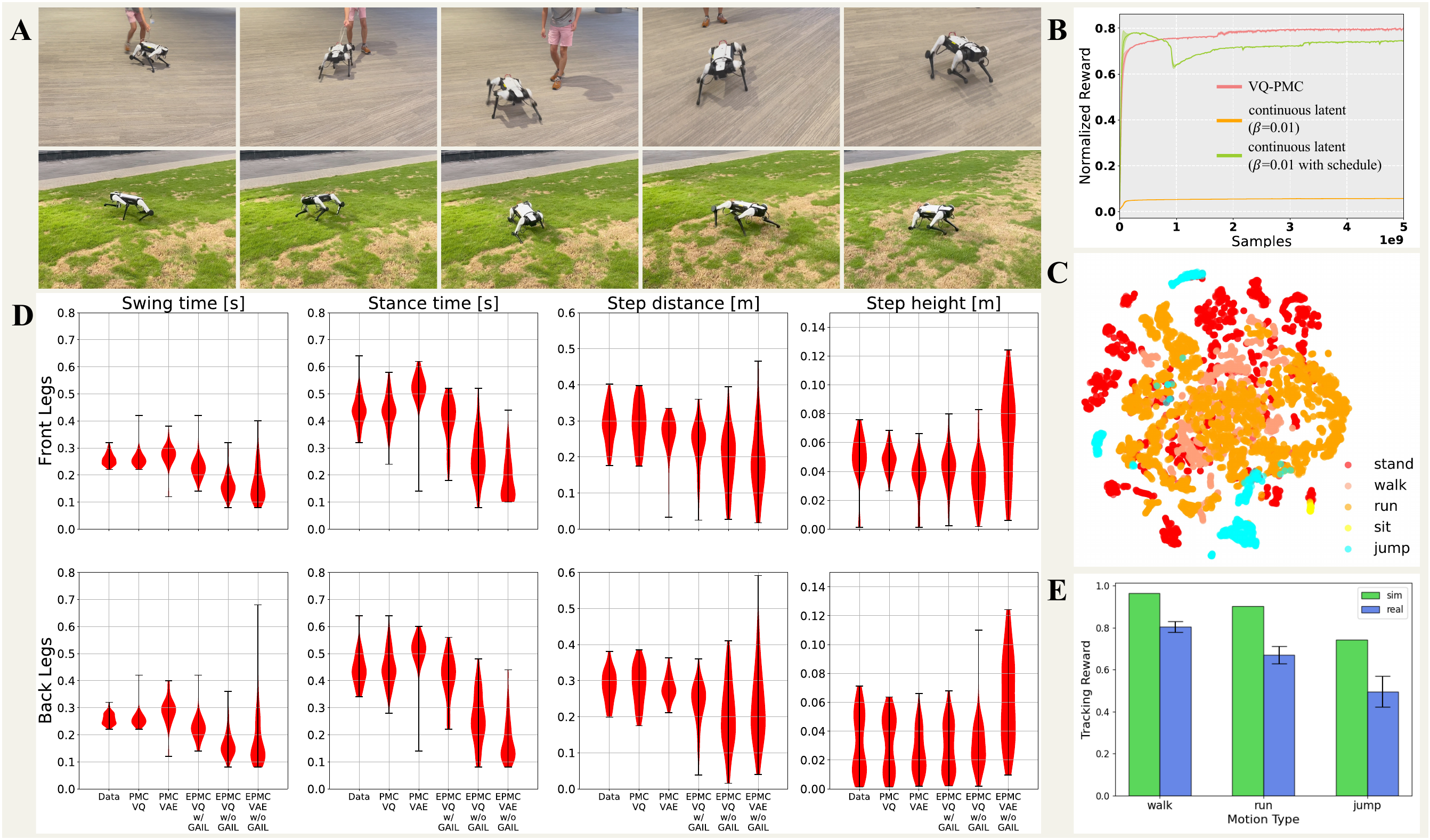}
\caption{\small Evaluation of the primitive motor controllers. (A) Snapshots of the MAX robot imitating motion data on different terrains. (B) Comparison of the learning curves of VQ-PMC, and $\beta$-VAE based methods in imitation learning. The experiments are repeated three times to plot the mean curve and the shaded region (standard deviation). (C) Visualization of the generated trajectories from the VQ-PMC network using t-SNE. (D) Gait analysis for the generated movements from the primitive motor controllers. The plots show statistics over an entire walking trajectory with $\sim$1000 frames/samples. The bands indicate the maximum and minimum values. (E) Comparison of the tracking rewards in simulation and reality. For each motion type, the experiments are repeated three times to compute the reward statistics in the real world, while the environment dynamics remain deterministic under this case in simulation and the reward is a deterministic value given the trained policy.}
\label{fig:pmc}
\end{figure}

This section evaluates the performance of primitive behaviours generated by the PMC. The PMC training is formulated as an RL problem by optimizing a tracking/imitation reward over all the animal motions. We use a single network to track all the animal motions in the dataset, where the motions are sampled with prioritized sampling by choosing motion clips that are hard to track in priority.

Figure~\ref{fig:pmc}A shows a few snapshots of the MAX robot in the real world when it is imitating a few motion clips in the animal dataset on different terrains. From these images and the Supplementary Videos, we observe that MAX shows lifelike animal movements and that the PMC is robust enough to deal with different terrains. Figure~\ref{fig:pmc}B evaluates the VQ-PMC method by comparing it with state-of-the-art approaches~\cite{bohez2022imitate,higgins2017beta} using Gaussian latent representations for motor control. Specifically, we compare with the $\beta$-VAE method~\cite{higgins2017beta} and its variant version with schedule annealing scheme introduced in ref.~\cite{bohez2022imitate}. These methods condition on a hyper-parameter $\beta$ that controls the closeness of the latent space with a prior normal Gaussian distribution. A larger $\beta$ will produce a learned latent space closer to the Gaussian prior but lower reproduced motion quality, whereas a small $\beta$ will lead to better imitation performance but less usable latent representations. Choosing an appropriate $\beta$ is difficult in $\beta$-VAE based method, and its variant with schedule annealing scheme can adaptively increase this parameter along the training period to guarantee the quality of the imitation performance and the latent representations. For the implementation of schedule annealing scheme, we follow the original setting in ref.~\cite{bohez2022imitate} that $\beta$ starts at 0 and anneals to a max value selected from $[0.001, 0.01, 0.1]$. In the comparisons, we find that the best value for maximum $\beta$ is 0.01. 
The training curves for the imitation learning tasks in simulation are depicted in Figure~\ref{fig:pmc}B. The normalized reward indicates the average reward of tracking all motion clips in the animal dataset with normalized scale in $[0,1]$.
As we can observe that VQ-PMC achieves the best tracking reward compared with the other baselines. $\beta$-VAE with schedule annealing reaches a comparable reward while it suffers from unstable training due to the manually designed annealing process. The $\beta$-VAE method with a constant $\beta$ fails to track these motion clips.

To understand the representation learned by VQ-PMC, we visualize the second last layer's output from the PMC network using t-distributed stochastic neighbour embedding (t-SNE)~\cite{van2008visualizing}. t-SNE is a dimensionality reduction technique that is well adopted in visualization of high-dimensional datasets. Figure~\ref{fig:pmc}C shows the scatter plot of the generated trajectories of the converged PMC in simulation. Each point in the plot represents a motion frame at a time step with its colour indicating its gait label. The gait is labelled manually from the dataset. As we can observe, different motion types can be understood by the learned representations, which can generally distinguish the motions in a low-dimensional space. The results demonstrate that the expressive capability of a deep generative model can be successfully incorporated into a motor control policy.

Next, we quantified how naturalistic the generated movements are from the PMC using gait analysis. Gait analysis is a well-established technique used in motion analysis~\cite{gouelle2018interpreting} and has a wide range of applications in veterinary medicine, for example, the diagnosis of lameness in dogs~\cite{jarvis2013kinematic, palya2022development}. 
Following ref.~\cite{palya2022development}, we recorded kinematic measurements of each joint and computed the spatiotemporal parameters, including swing time, stance time, step distance and step height of the learned behaviours. The swing time is the period that a foot is not in contact with the ground; the stance time is the period when the foot is in contact with the ground; the step distance is the distance of the given limb's heel strike from the opposite limb's heel strike in the direction of movement; and the step height is the peak of the given limb's toe height for each swing time. For more details please refer to ref.~\cite{palya2022development}. 
To achieve fair evaluation, we record these kinematic parameters on flat terrain in simulation and set up four comparison groups. The control group containing all the motions from the animal dataset on flat terrain depicts the groundtruth of naturalness. The second and third groups collect movements from VQ-PMC and $\beta$-VAE with an annealing schedule by imitating these animal motions, respectively. We also include the evaluation of EPMC controllers to see whether reusing the primitive-level network to solve environmental-level tasks will lose naturalness. The average speed from the animal motion data is around 0.8m/s, so we set the target speed as the same for the EPMC contollers. Therefore, the fourth and fifth groups contain the movements gathered from the learned EPMC on flat terrain following the navigation command, where the target direction in horizontal plane is sampled uniformly from 0 to 360 degrees every 5 seconds, which is roughly consistent with the walking behaviours in the animal motion data. The difference is that the EPMC controller in the fourth group is trained with an additional adversarial imitation reward (as will be detailed in Section~\ref{sec:method}).
The final group contains the movements generated from the learned EPMC based on the primitive-level network trained from $\beta$-VAE with an annealing schedule.
Figure~\ref{fig:pmc}D summarizes the gait analysis over the generated behaviours in simulation, visualized using a violin plot.
The rows represent the front and hind legs, and each column represents one spatiotemporal parameter. The aforementioned groups are denoted as data, PMC VQ, PMC VAE, EPMC VQ w/ GAIL, EPMC VQ w/o GAIL and EPMC VAE, respectively. In each panel, the red region indicates the distribution of the corresponding metric, and the top and bottom lines represent the extremes. We observe that the VQ-PMC can accurately reproduce the movements in the animal motion data that the shape of the violin plot of VQ-PMC matches that of data much better than $\beta$-VAE with annealing schedule.
The remaining three groups have to follow a random command and preserve animal naturalness simultaneously, and therefore the motion naturalness shows degradation. However, VQ-based EPMC still shows closer matching with the motion data compared with $\beta$-VAE based EPMC, and the naturalness of EPMC can be improved by applying an additional adversarial imitation reward. 

Finally, Figure~\ref{fig:pmc}E reports the performance gap between simulation and reality when applying the trained VQ-PMC in reality via a zero-shot manner. The green and blue bars show the average normalized reward in simulation and reality, respectively. Each trajectory-tracking episode is repeated three times in real-world experiments to compute the average reward and the variance. As expected, the rewards decrease when we directly apply the trained controller on real robot considering the existence of the dynamics gap between simulation and reality. At the end of Supplementary Video~1, we observe that the imitation behaviour well matches that in simulator despite the existence of the gap.

\subsection{Traversing Complex Obstacles}

\begin{figure}[t] 
\centering
\includegraphics[width=\linewidth]{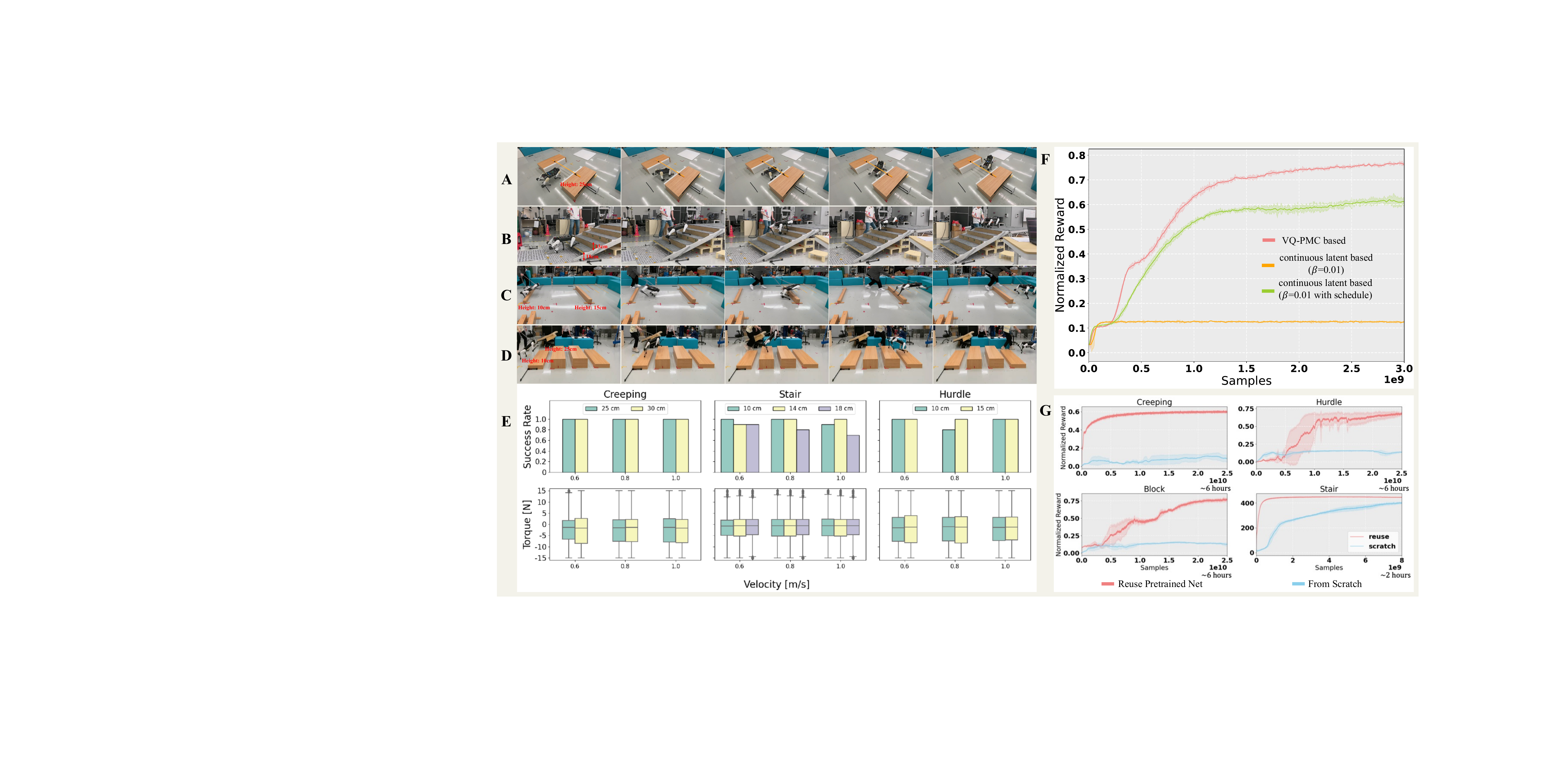}
\caption{\small Performance evaluation of the environmental-primitive motor controllers. (A-D) Snapshots for creeping (A), ascending stairs (B), jumping over hurdles (C) and freerunning over blocks (D). 
(E) Success rate and output torque distribution for three elementary tasks in real-world experiments. Each elementary task is configured with a single corresponding obstacle. Each elementary experiment is repeated for 10 times for success rate statistics. The torque distribution is computed from all samples of the 10 repetitions.
(F) Comparison of the effectiveness of reusing different pre-trained primitive-level networks learned by VQ-PMC and $\beta$-VAE based methods. The curves indicate the training of the environmental-level network on flat terrain. The experiments are repeated three times to plot the mean curve and the shaded region (standard deviation).
(G) Comparison of the learning curves for different EPMC controllers by reusing the proposed primitive-level networks and training from scratch. All the experiments are repeated three times to plot the mean curve and the shaded region indicating the standard deviation.}
\label{fig:epmc}
\end{figure}

We evaluate the performance of the EPMC controllers on solving various traversing tasks. In real-world experiments, for ascending stairs, we built stairs with five-level steps where the first step is of height 18cm and the upper four steps are 13cm high. For the creeping task, we used a pole spanning two platforms at 25cm to create a narrow path for the robot. For jumping over hurdles, we created two types of hurdles of sizes 110cm $\times$ 10cm $\times$ 10cm and 110cm $\times$ 10cm $\times$ 15cm for length $\times$ width $\times$ height.
Freerunning over blocks consisted of four consecutive blocks with two sizes of 
110cm $\times$ 50cm $\times$ 10cm and 110cm $\times$ 50cm $\times$ 25cm for length $\times$ width $\times$ height, respectively.
Figures~\ref{fig:epmc}A-D show these scenarios in real environments. In the scenarios depicted, the MAX robot traverses over the stairs, the spanning pole, the hurdles and the blocks, respectively. The recording in Supplementary Video~2 demonstrates that the MAX robot exhibits lifelike animal movements. Note that the behaviours generated by the neural network in these tasks are not captured in the animal motion data we use, implying that the trained policy understands well the naturalness of animal motion and generalizes the knowledge to adapt to new tasks. To evaluate how robust these capabilities are in real-world scenarios, we test the trained controllers in three elementary sub-tasks from these scenarios by varying the heights of the creeping path, the stair and the hurdle, respectively. The top charts in Figure~\ref{fig:epmc}E show the average success rate from 10 repetitions of each sub-task under variable heights and command speeds in real-world experiments, and the bottom charts show the average output torques from these sub-tasks. As we can observe, the trained controllers maintain high success rates under various settings, and the output torques are distributed around relatively small absolute values even for these agile behaviours. In simulation, the success rate is always perfect (100\%) because the reward and the convergence conditions guarantee this. Supplementary Video~4 provides a quick summary of the performance of the converged policies in simulation.

At the imitation learning stage in simulation, we have compared the proposed VQ-PMC method with $\beta$-VAE based motor controllers. At the current stage, we continue to investigate the quality of the primitive-level networks trained by VQ-PMC and $\beta$-VAE based methods when reusing them to solve environmental tasks. 
All the experiments are repeated for three times to plot the average value and the variance region.
Figure~\ref{fig:epmc}F compares the learning curves of reusing various pre-trained networks for training the environmental-level network on flat terrain following a command. As clearly shown, the primitive-level network trained by VQ-PMC significantly accelerates the training of the EPMC controllers compared to other baselines. Moreover, we conduct ablation studies to compare our method with a baseline method which shares the same neural network architecture as ours while it is randomly initialized without reusing pre-trained primitive-level network. The training curves for these comparisons in four traversing tasks are depicted in Figure~\ref{fig:epmc}G. 
As we can clearly observe, reusing pre-trained primitive motor priors significantly facilitates the training of the EPMC controllers, while training from scratch is slow or even unable to solve the tasks, considering the difficulty of the tasks and sparse rewards used therein (as will be detailed in Section~\ref{sec:method}).
The policies can solve the tasks within 6 hours (with nearly perfect success in simulation), but to obtain solid and robust pre-trained models, we ran the training tasks for a longer time (2–4 days), to obtain a final stable checkpoint.

\begin{figure}[h] 
\centering
\includegraphics[width=.72\linewidth]{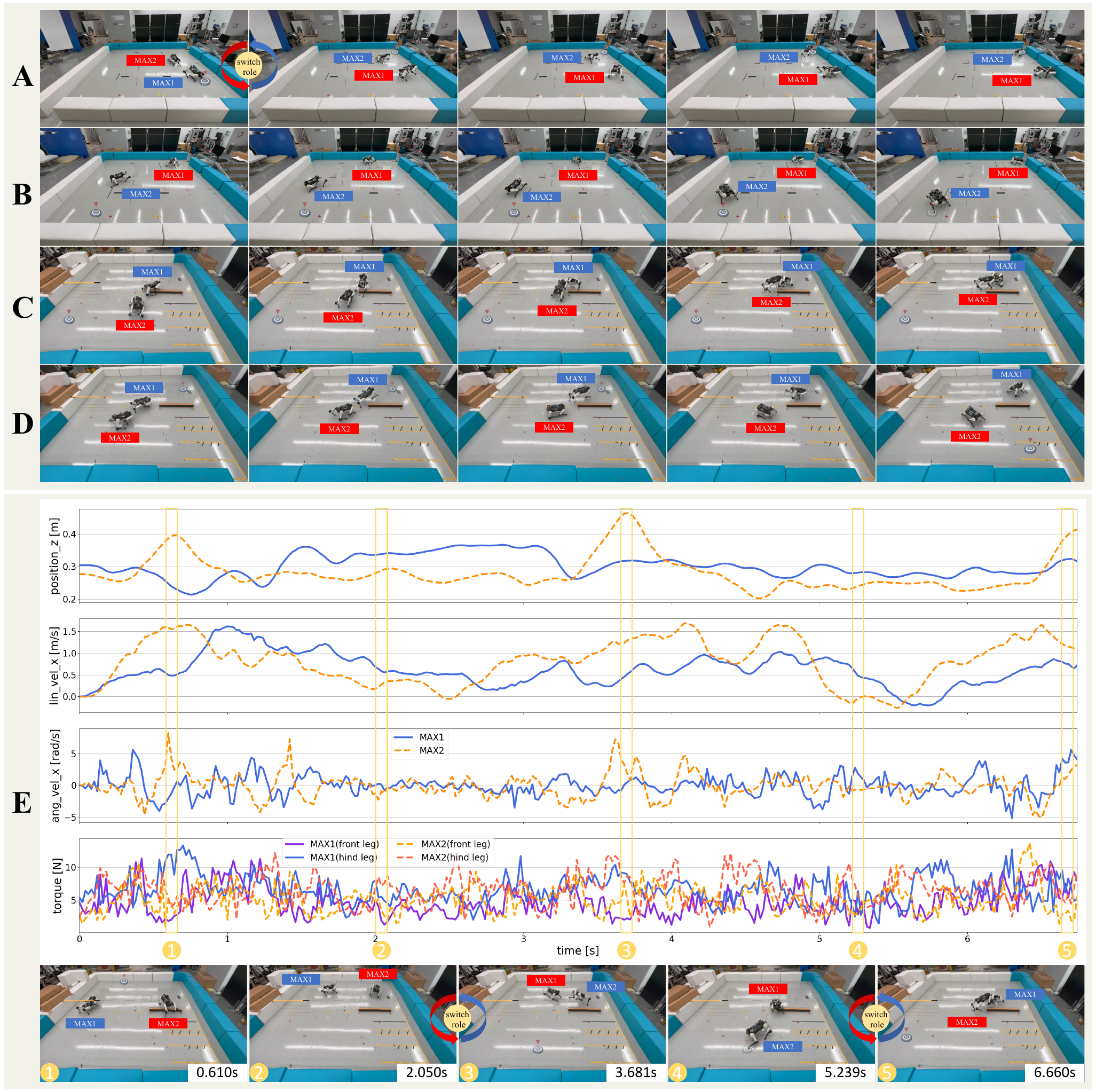}
\caption{\small
Snapshots in the Chase Tag Game. (A) A case in which the chaser, MAX2, gives up chasing MAX1 when MAX2 estimates that it is not possible to catch MAX1 before MAX1 reaches the flag. (B) 
A case in which the chaser, MAX1, hesitates and wanders around. (C) A case in which the chaser, MAX2, pounces on MAX1. (D) A case in which the evader, MAX1, pounces on the flag.
(E) A detailed analysis of the torques, angular velocities, linear velocities and root heights of the two robots in a complete game episode.
} 
\label{fig:sepmc}
\end{figure}

\subsection{Chase Tag Game}

We evaluate the ability of the learned SEPMC to solve complex downstream task in this subsection. Specifically, we design a challenging and interesting setting that is a multi-agent game, in which two MAX robots play against each other to alternatively fill the roles of chaser and evader, where the roles are determined by a flag scattered in the playground. The game setting is similar to that in World Chase Tag~\cite{wct}, an international championship for competitive human parkour, with simplification and novel elements. The game takes place in a 4.5m $\times$ 4.5m square equipped with hurdles and a spanning pole. It is not easy to involve larger obstacles within the limited area size. At the beginning of each game episode, two MAX robots are randomly scattered in the play area, and an additional flag is randomly placed as well. The roles of the robots are initially assigned randomly, where one MAX robot is the chaser and another is the evader. When the game starts, the mission of the chaser is to chase the evader, and as long as the distance between the two MAX robots is shorter than 0.6m, the game is terminated with the current chaser winning the game. Meanwhile, the evader aims to avoid being chased, while it has the chance to switch the roles of the current players once it reaches 0.3m from the flag. If this happens, the evader turns into the chaser while the previous chaser becomes the evader; the flag disappears and re-spawns randomly in the area. The game proceeds until the case becomes true that the distance between the two robots is smaller than 0.6m. 

A simulation environment implementing the aforementioned setting is used for training the SEPMC. In simulation, an agent only receives a non-zero reward at the end of one episode, where the winner receives a sparse reward of $+1$ while the loser receives $-1$ to satisfy a zero-sum two-player game. For solving the multi-agent RL problem, we adopted prioritized fictitious self-play (PFSP)~\cite{vinyals2019grandmaster,han2020tstarbot}. 
Detailed settings of the PFSP procedure are identical to those used in ref.~\cite{zhu2023ncp}.
The training takes approximately 40 hours to reach a convergent SEPMC by reusing the pre-trained environmental- and primitive-level networks, given only a sparse win–loss binary reward. We stored the network parameters each hour as the training proceeds. In total, 40 models were stored, and we evaluated them with a round robin tournament~\cite{elo}. 
Each pair of players were evaluated with 100 matches for computing the Elo scores~\cite{elo} (a common metric for measuring the overall performance of players in a round robin tournament). The Elo scores are reported in Supplementary Section 6.1.3.
The result demonstrates that the trained SEPMC can consistently improve over historical models under the chase tag game setting.

In the real robot experiments, we let the strategic-level network output only the navigation direction to the environmental-level network, and we assigned the robot a command on the desired speed manually to control the dynamics. Specifically, we set the root linear velocity to 0.5m/s (the lower bound of the speed command range in training) for real robot experiments due to the limited area size. A faster speed would have led to unsafe situations in this small area and affected the game experience.

The real robot experiments are shown in Supplementary Video~3.
We observed the robots employing lifelike strategies in the games. For example, the chaser will give up chasing if the chaser estimates that there is no chance to catch the evader before the evader reaches the flag (Figure~\ref{fig:sepmc}A and Supplementary Video~3); 
at this time, the chaser will hesitate and wander around, waiting for the re-spawn of the flag (Figure~\ref{fig:sepmc}B and Supplementary Video~3). 
Moreover, at the very moment that the chaser is about to catch the evader, the chaser prefers to perform a swift pounce, like an animal pouncing on its prey (Figure~\ref{fig:sepmc}C and Video~3); 
and similar behaviours were observed when the evader was about to reach the flag (Figure~\ref{fig:sepmc}D and Video~3).
The markers in Figure~\ref{fig:sepmc} indicate the game roles and the inherent identities of the two robots: `MAX1' and `MAX2'. The red-coloured rectangle indicates the chaser in this frame, and the blue-coloured rectangle indicates the evader.

In Figure~\ref{fig:sepmc}E, we plot the analysis of one complete episode of the real-world game, where the curves show the torque of the actuators, the angular velocities, linear velocities and the root heights of the two MAX robots. The frames below the curves are the according frames aligned by the indices 1–5 for five different time steps. For example, in the first frame in Figure~\ref{fig:sepmc}E, MAX1 creeps through the spanning pole, and its actuators output relatively large torque for its hind legs for supporting its body. Overall, the torques fall into reasonable intervals during the entire game for both MAX robots.

Lastly, we were curious about how much a human-operated conventional controller could score when playing against the trained neural controllers in this game. To achieve this, one of our researchers used a joystick to control the MAX robot via the XMPC algorithm~\cite{chi2022linearization}, while the other MAX robot was controlled by the trained SEPMC. The maximum root linear velocity of the XMPC controller was set to 0.8m/s, which is faster than that of the trained model, to complement the lower flexibility of the conventional controller compared to the trained model with agile behaviours. Conventional controllers usually output periodic gait control signals, like trotting. Therefore, they can not perform movements like sharp turning or pouncing. The human-operated robot lost two games 0:2 against the one with the trained model.
These results demonstrate that the MAX robot using XMPC methods shows less flexibility, and it is not easy to control the robot manually in the chase tag game, which requires agile movements and strategy. The games are shown in Supplementary Video~3.

\section{DISCUSSION}
\label{sec:discussion}
We have proposed a general learning framework for reusing pre-trained knowledge at different perception and task levels. To produce lifelike movements, we have proposed a novel control method for learning expressive discrete latent representations, benefiting from the power of deep generative VQ-VAE models. The discrete representations are effective in facilitating training of the environmental-level network. For the environmental-level learning stage, we have solved a number of challenging traversing tasks that are rarely considered by previous approaches, including creeping across spanning poles, and jumping over hurdles and blocks. For reusing these movement capabilities in solving downstream tasks, we have designed a challenging chase tag game. Using self-play and multi-agent RL, the final complete controller can perform lifelike strategies and behaviours. We have successfully applied all these trained controllers on the MAX robots, on which animal-level strategies and movements emerge in real-world tasks. It is also worth mentioning that when dealing with new tasks that require specific movement skills not considered in the present environmental-level network, we only need to train an additional model on that skill and compress the ability into the uniform environmental-level network using distillation method. For example, in Supplementary Video~5, we also involve a fall recovery model as considered in previous learning-based control approaches~\cite{hwangbo2019learning,yang2020multi}. 
The training details of the fall recovery are provided in Supplementary Section~6.1.2. All the network structures are detailed in Supplementary Section~6.3.

In our main results, the MAX robot is not equipped with onboard cameras and relies on a motion capture system to acquire the terrain information. To further demonstrate the generality and scalability of our framework, we show that replacing the motion capture system with onboard sensing is convenient. Similar to refs.~\cite{doi:10.1126/scirobotics.abc5986,cheng2023extreme,zhuang2023robot}, we take an additional student–teacher distillation procedure, where our original trained environmental-level network is treated as the teacher, and we construct a student environmental-level network that takes only depth image from the onboard camera as input, in addition to the proprioception and navigation command. The student environmental-level network also outputs a code to drive the primitive-level network, which reuses the same frozen pre-trained PMC model. So, we only need to perform a supervised learning/distillation procedure to train the student environmental-level network given the teacher environmental-level network's output, that is, the latent code, as labels. Then, we installed an Intel RealSense Depth Camera D455 at the head of the real MAX robot. Supplementary Video~6 demonstrates that the MAX robot with the onboard camera reproduces lifelike agility. The training details of the student network are provided in Supplementary Section~6.4.

The proposed framework shares some similarities with existing methods~\cite{bohez2022imitate,liu2022motor,xie2020learning}, which also adopt multi-stage learning schemes. Although ref.~\cite{liu2022motor} proposes reuse of pre-trained primitive skills, these skills, like shooting and dribbling, are specific for the simulated football game. Moreover, the proposed vector-quantized primitive motor controller has been shown to be more expressive than Gaussian-based latent representations adopted in ref.~\cite{liu2022motor} for generating lifelike behaviors. 
The methods in refs.~\cite{xie2020learning,bohez2022imitate} also aim to pre-train a primitive controller that imitates animal motion data, while the generated performance from the real robots only covers general locomotion movements. Our considered traversing tasks at the environmental-level and in the chase tag game are much more challenging compared with these existing approaches.

During the review process of the current work, we identified a few concurrent approaches~\cite{cheng2023extreme,zhuang2023robot,hoeller2023anymal,caluwaerts2023barkour} that also aim to learn agile traversing skills using end-to-end neural networks. The skill training pipelines of these methods share some similarities that carefully hand-designed rewards and task curriculum are necessary to obtain the final policies. The major differences between these methods and ours are that (1) we embrace the concept of pre-training and organizing the knowledge into hierarchies according to perception and control levels; (2) we propose pre-training primitive behaviours from animal motions using a generative model. The generated movements are naturally lifelike and energy-efficient, and we do not need to focus on reward engineering; (3) reusing bottom pre-trained models significantly facilitates the learning of higher hierarchies, without the need for a task curriculum; (4) our approach is the only work considering strategic-level training and experimenting with multi-agent games. We perform a detailed discussion and comparison with these state-of-the-art methods in Supplementary Section~6.5.

One future research direction is that for data-driven methods the acquisition of motion data is expensive and more economical data collection ways are in demand. An alternative method could extract necessary motion information from online video resources, and several works have demonstrated impressive results for humanoid characters in simulation~\cite{peng2018sfv, zhang2023vid2player3d}. Transferring such techniques into real robot control promises to be a potential direction.

\section{METHODS}
\label{sec:method}

\subsection{Primitive-Level Training}
In this section, we provide details for the learning stage of shaping the primitive-level knowledge. The PMC is trained by imitating collected animal motion clips.

\noindent
\textbf{Motion Data Acquisition and Retargeting.} The motion data is collected from a motion capture system. We invited a trained medium-sized Labrador retriever to perform natural movements, and we captured sequences of locomotion of various gaits and root trajectories. 
The Labrador was guided by a dog trainer to follow various instructions. The gaits included walking, running, jumping and sitting. Each gait was repeated four to six times to ensure data diversity. 
The Labrador was guided to follow straight, square and circle trajectories. We obtained approximately half an hour of motion data captured at 120 fps. 
In addition, the Labrador was guided to ascend and descend a set of stairs, which had three steps with a fixed step width of 0.32m and height of 0.16m. The data with respect to stairs contains about 9 minutes of motions captured at 120 fps. In the tracking problem of PMC training, for the motion clips involving stairs, we recreated a set of simulated stairs of the exact same size to precisely match the stairs in the real-world motion collection, since the mission of the PMC is to precisely track the motion clips. The generalization to various shaped stairs was deferred to the EPMC training stage. We only collected data to cover the above-mentioned movements, and the scenarios reported in our EPMC experiments (like creeping and jumping over obstacles) were not considered during data collection phase. The proposed method can learn from these motions and generalize to various scenarios in EPMC training that are absent during the motion capture phase.

Since the morphology of the animal is different from that of the robot, we retargeted the source motion of the animal to the legged robot using inverse kinematics~\cite{gleicher1998retargetting}. The retargeting was implemented by first selecting a set of key points, including shoulder, shoulder blade and haunch from the animal, and these key points were used to compute the position and orientation of the robot base. Then, the fore toes and hind toes from the animal were aligned with the corresponding end effectors from the robot. After the pose of the root and end effectors were determined, the rest of the robot's joints were solved using inverse kinematics following ref.~\cite{RoboImitationPeng20}. 
To further reduce the morphology discrepancy, we lowered the robot's root height and enlarged the distance between the legs on the left-hand side and right-hand side, resulting in more natural-looking behaviour on the MAX robot. All the source motion data and retargeted data are released in this paper.

\noindent
\textbf{Problem Setup}
We use RL to solve the imitation problem. RL solves an online decision problem where the dynamics in the system is described by a Markov Decision Process (MDP). 
At time step $t$, the agent performs an action $a_t$ conditioned on a state $s_t$, and then receives a reward $r_t$ and the next state $s_{t+1}$ from the environment.
Let $p\left(s_{t+1} | s_t, a_t\right)$ be the state transition probability. The objective of RL is to learn a policy $\pi(a_t | s_t)$ that maximizes the expected cumulative reward $R(\tau)$ over trajectories
\[
J_{\text{RL}}(\pi)=E_{\tau \sim \pi(\tau)}[R(\tau)]=E_{\tau \sim \pi(\tau)}\left[\sum_{t=0} \gamma^t r_{t}\right],
\]
where $\pi(\tau)=p\left(s_0\right) \prod_{t=0} \pi\left(a_t | s_t\right) p\left(s_{t+1} | s_t, a_t\right)$ 
denotes the probability of a trajectory $\tau$; $p\left(s_0\right)$ is some distribution of the initial state; 
and $\gamma \in [0, 1]$ is a discount factor. 

We aim to transfer the expressive power of large deep generative models into
motion generation for robot control. Specifically, we consider Vector Quantized Variational AutoEncoders (VQ-VAE)~\cite{van2017neural}, a deep representation learning method that demonstrates surprising performance on image generation~\cite{ramesh2021zero}.
VQ-VAE combines the strengths of Variational AutoEncoders (VAEs) and Vector Quantization (VQ) to learn discrete latent embeddings. 
In VQ-VAE, the encoder maps the input to one latent embedding and the decoder maps the embedding back into an output to recover the input. The VQ-VAE training loss consists of a reconstruction loss and a commitment loss as denoted below
\[
    L_{\text{VQ-VAE}}=\log p\left(x | z^q(x)\right)+\left\|sg\left[z^e(x)\right]-e\right\|_2^2+\beta\left\|z^e(x)-{sg}[e]\right\|_2^2,
\]
where $\log p\left(x |z^q(x)\right)$ is the reconstruction loss for the input data $x$; 
$z^q(x)$ is the quantized vector; 
\emph{sg} is an operator to stop the gradients;
$z^e(x)$ is the output of the encoder;
$e$ is the nearest embedding from the discrete latent embeddings to $z^e(x)$;
and the hyperparameter $\beta$ balances the last two terms.
In our problem, distinct from VQ-VAE which reconstructs input samples unconditionally, the PMC generates control signals conditioned on certain robot proprioceptive states.

\noindent
\textbf{Vector Quantized Primitive Motor Controller.} 
In PMC, we devise a conditional encoder-decoder structure that expands the VQ-VAE structure for robot control. As shown in Figure~\ref{fig:overview},
the MLP encoder $p^{e}(z_t^e | s_t^p,s_{t+1:t+k}^f)$ takes the proprioception $s_t^p$ and the target future trajectory $s_{t+1:t+k}^f$ from the animal motion data as input, and it maps them to a latent variable $z_t^e$. 
On the other hand, the discrete latent embeddings are represented as $\mathbf{e}\in R^{K \times D}$, where $K$ is the number of latent embeddings, and $D$ is the dimensionality of each latent embedding vector $e_i$ for $i \in 1,2, \ldots, K$. 
Then, the encoder output $z_t^e$ is compared with all the embeddings in $\mathbf{e}$, and the one with the nearest distance to $z_t^e$ is selected as the quantized vector $z_t^q$. This vector quantization process can be formulated as
\[
z_t^q=e_k, \ \text{where}\ k=\arg\min_{j}\left\|z_t^e-e_j\right\|_2.
\]

The decoder $\pi^{d}(a_t | s_t^p,z_t^q)$ takes the state $s_t^p$ and the embedding $z_t^q$ as input, and outputs $a_t$ as the primitive motor action, which specifies the residual target joint positions for proportional-derivative (PD) controller. 
That is, the target joint positions are calculated by adding the action $a_t$ to the current joint positions. The PMC is queried at 50Hz and the control frequency of the PD controller is 500Hz.

The proprioceptive state of the robot $s_{t}^p$ at time step $t$ is the concatenation of three consecutive states obtained from time steps $t-2$ to $t$ and historical actions from time steps $t-3$ to $t-1$. 
A single state at each time step consists of joint positions, joint velocities, root linear velocity, root angular velocity and gravity vector of the robot. Root linear and angular velocities are calculated in the robot coordinate system. Compared to directly output joint torques, PD controllers have been demonstrated stable for learning-based control methods~\cite{peng2017learning}, and we follow this setting in this approach.
The future trajectory $s_{t}^f$ at time step $t$ specifies the target pose for the robot. 
It consists of states of reference motion spanning over 0.03s, 0.06s, 0.3s and 1s in the future. 
Each reference state includes the reference joint positions, reference root position and orientation that calculated in the current robot coordinate system. The PMC is trained to imitate all the animal motion data with the tracking reward defined similar to ref.~\cite{RoboImitationPeng20}. The detailed reward function is provided in Supplementary Section 6.1.1.

\noindent
\textbf{Prioritized Sampling.}
We propose to imitate all motion trajectories by a single PMC. Similar to the data imbalance issue commonly encountered in standard supervised learning, some motion clips that are rare from the dataset might be underfit in the imitation policy. For example, jumping movements are captured less frequently compared with common walking behaviours, and these agile movements are usually even more challenging to imitate than others. The consequence is that the imitation policy fails to reproduce such agile movements. Instead of uniformly sampling motion trajectories in the dataset, we utilize prioritized sampling, where the motion clip $m_i$ from the dataset $M$ is sampled with probability
\[
p_i=\frac{f(R_{m_i})}{\sum_{m \in M} f(R_m)},
\]
where $R_m\in[0,1]$ are the cumulative rewards obtained for motion clip $m_i$, and
$f(x)=(1-x)^{\alpha_1}$ such that the method samples motion clips that are hard to imitate, where $\alpha_1 \in R_{+}$. 

\subsection{Environmental-Level Training}

After training the PMC, we removed its encoder and inherited the trained discrete latent embeddings and the decoder. These pre-trained neural network parameters are fixed, and we created a new environmental-level network on top of it to output a categorical distribution over the discrete latent embeddings, as shown in Figure~\ref{fig:overview}. 
At the previous learning stage, the controller is blind without knowing any exteroceptive information. Therefore, the PMC extracts sufficient lifelike latent representations, while it is not capable of connecting these lifelike movements with the environment. For the current stage, we aim to align its knowledge in the pre-trained primitive-level network to adapt to various challenging environments. In addition to the proprioception, the environmental-level network takes the exteroception $s_{t}^{ex}$ to simulate the perception of LIDAR and depth cameras. $s_{t}^{ex}$ includes a $25\times 13$ map to sense the terrain height within a square area of $2.4 \mbox{m} \times 1.2 \mbox{m}$ around the robot, and a $25\times 13$ map to sense front depth of a region of $0.5 \mbox{m} \times 0.6 \mbox{m} \times 3 \mbox{m}$ for width $\times$ height $\times$ depth. 
Additionally, 128 rays evenly splitting across 360 degrees are horizontally emitted from the centre of the robot to sense the distance from the robot to the surrounding environment.

Moreover, we propose to control the EPMC and let it receive an additional outer command on determining the direction and speed of the robot by adding a command compliance reward to the traversing task reward. The command is randomized with arbitrary direction and a target random root linear velocity ranging from $[0.5, 3.0]$ m/s periodically. More details on the reward setting can be found in Supplementary Section~6.1.2.

\noindent
\textbf{Training Environments.}
In addition to general locomotion on flat terrains and stairs, we considered a number of new challenging traversing tasks, including creeping, jumping over consecutive hurdles and freerunning over consecutive blocks with lifelike behaviours. These tasks are rarely considered in existing approaches. Below, we provide details for the training procedure for these traversing tasks.

\noindent
\textbf{Generative Adversarial Learning for Flat Terrain Locomotion.}
For general locomotion on flat terrain, because the motion data were also collected on flat terrain, the EPMC is required to preserve the naturalistic movements strictly and follow the command meanwhile. To achieve this, the task reward is defined as an adversarial imitation reward plus a command compliance reward. Note that at this stage the controller will not condition on reference trajectories from the motion data anymore.

Generative adversarial imitation learning (GAIL)~\cite{ho2016generative} is initially proposed to imitate expert trajectories by minimizing the distance between distributions of the generated trajectories from the policy and those of the expert. GAIL introduces a trainable discriminator to distinguish the two distributions, and the policy is optimized to fool the discriminator to generate data that is hard to be distinguished. Following GAIL, we applied a discriminator $D(s^p,a)$ to classify if a given state-action pair $(s^p,a)$ is produced from the previous imitation training stage by the PMC or the current training stage by the EPMC. The objective for the discriminator is
\[
\min_{D} -E_{(s^p,a)\sim \pi^{\text{PMC}}}[\log D(s^p,a)]-E_{(s^p,a)\sim\pi^{\text{EPMC}}}[\log(1-D(s^p,a))],
\]
where $\pi^{\text{PMC}}$ indicates the PMC policy, and $\pi^{\text{EPMC}}$ denotes the EPMC policy. The imitation reward is defined as $r_{\text{GAIL}}=-\log(1-D(s^p,a))$.
The total reward is the weighted sum of the command compliance reward and the imitation reward. Please refer to Supplementary Section 6.1.2 for the detailed reward function.

The GAIL reward is not applied to other traversing tasks except on flat terrain, since the motion data do not contain natural animal motions in the exact same scenarios as considered by other tasks.

\noindent
\textbf{Residual Control for Traversing Stairs.}
Different from previous approaches~\cite{agarwal2023legged,doi:10.1126/scirobotics.abc5986} which also consider the task of traversing stairs, we propose the reuse of the primitive-level network to go over stairs with lifelike animal behaviour. For training an EPMC in this scenario, the environmental-level network has to adapt to various shaped stairs with naturalistic movements like animals. To achieve this more efficiently, the environmental-level network for solving this task outputs an additional residual control offset to the primitive motor action, and the final motor action is the sum of the primitive motor action plus the offset~\cite{li2023learning}. The residual control helps the entire controller to adapt to different shaped stairs faster while still maintains naturalistic movements, while we find that this technique is not necessary for other traversing tasks.

\noindent
\textbf{Simple Reward Structure.}
In previous locomotion tasks on flat terrain and stairs, the command compliance reward was applied at each time step to constrain the agent to follow the command immediately. However, for the tasks of creeping, and running over hurdles and blocks, the same reward will prevent the agent from succeeding. This is reasonable given that the considered tasks here are sensitive to the robot's speed. For example, it is not practical to creep too fast, while it is not possible to jump over hurdles and blocks with slow movements either. Constraining the agent at each time step to follow a given speed is not appropriate in such cases. Instead, we propose to apply these hard traversing tasks a sparse reward to only count the average speed at the end of the episode. That is, the agent receives a speed compliance reward (which is always positive) only when it solves this task. Using such rewards, the agent has to solve the task in advance at least and then adapt to a desired speed. The navigation direction rewards follow the same setting as locomotion on flat terrain. Although sparse rewards put additional burden for exploration, we observe that the agent could adapt to solve these tasks by adjusting its speed dynamically during the training progress, thanks to the powerful pre-trained primitive-level network. Note that many existing approaches on learning various locomotion skills~\cite{hwangbo2019learning,doi:10.1126/scirobotics.abc5986, doi:10.1126/scirobotics.abk2822, kumar2021rma,cheng2023extreme,zhuang2023robot,hoeller2023anymal,yang2023cajun,caluwaerts2023barkour} paid considerable efforts on designing a complex reward function, including a considerable number of (even more than ten) items focusing on the naturalness, energy-efficiency, gait pattern and navigation compliance and so on. Moreover, a specific curriculum scheme is often involved as well. Different from these approaches, our PMC pre-trained on animal dataset naturally generates lifelike and energy-efficient movements, and we do not need to focus on carefully hand-designed rewards and curriculum for the above consideration. Our reward function for skill learning (EPMCs) only contains two terms on the navigation command (Supplementary Section 6.1.2).

\noindent
\textbf{Multi-Expert Distillation.} 
We use the policy distillation method~\cite{rusu2015policy} to compress all the learned environmental-level networks into a uniform one. The distillation loss function is defined as follows
\[
    \min_{\bar{\pi}^{\text{EPMC}}}\ 
    E_{i\sim\{\text{flat,creeping, hurdle, block, stair,} \cdots\}}
    E_{\tau\sim\text{env}_i,\bar{\pi}^{\text{EPMC}}}
    KL\left(\bar{\pi}^{\text{EPMC}} || \pi^{\text{EPMC}_i} \right),
\]
where $i$ indicates the index of a scenario, $\bar{\pi}^{EPMC}$ denotes the uniform EPMC,
the trajectory $\tau$ is sampled from the uniform EPMC with a uniformly sampled environment $\text{env}_i$ from the considered traversing tasks, and the teacher environmental-level network $\pi^{\text{EPMC}_i}$ is selected according to the environment. 
Note that for the stair task, the residual control is adopted, and this structure is kept in the uniform EPMC for matching the teacher's output; while for other tasks where residual control is not used, the residual part in the uniform EPMC simply predicts zero offsets for the motor actions. When we need to deal with new tasks that are not considered in the existing environmental-level networks, we only need to train an additional environmental-level network on that skill and compress the ability into the uniform environmental-level network using the above method.

\subsection{Strategic-Level Training}

The SEPMC solves complex tasks by reusing the primitive- and environmental-level networks to ensure motion quality and learning efficacy. The inputs to the strategic-level network consist of task relevant information, proprioception and exteroception, and the strategic-level network outputs the navigation command to the environmental-level network. In the chase tag game, the task-relevant information contains the current role of the robot (a binary value indicating chaser or evader), the information on its opponent robot (including the opponent's global position, orientation, root linear velocity and root angular velocity), and the global locations of itself and the flag.

We employed PFSP, an effective multi-agent RL training scheme that can produce strong AI agents~\cite{vinyals2019grandmaster,han2020tstarbot}.
In PFSP, we sample opponents for the current player using the same prioritized sampling strategies as proposed previously~\cite{vinyals2019grandmaster,han2020tstarbot}, where the current player chooses its most challenging opponents in priority from the historical models. That is, at each training step, multiple concurrent games are ongoing, and in each of these games the current training agent is playing against one of its historically stored versions in probabilities. For every hour, we dump the parameters of the entire SEPMC policy and keep it as a historical version of the agent in a candidate set. Once a new game starts, the current training agent chooses its opponent from the candidate set following the probability
\[
p_i=\frac{g(P_i)}{\sum_{o\in O}g(P_o)},
\]
where $g(x)=(1-x)^{\alpha_2}$ is a weighting function, $P_o\in[0,1]$ is the probability that the current training agent wins the opponent $o$, and $\alpha_2$ is a hyperparameter.

\subsection{Transferring to Reality}
To facilitate robust zero-shot transfer, we follow most existing learning-based robot control approaches~\cite{tan2018sim,xie2020learning,doi:10.1126/scirobotics.abc5986,bohez2022imitate} to randomize the terrain friction, the actuator torque limit, and apply disturbances with random forces on the root of the robot, periodically. Additionally, for the obstacles considered in our tasks, we add inobservable inflated cylinders with random radius to cover the obstacle edges. This trick encourages the agent to step far away from the edges to reduce the falling risk. Details of the domain randomization parameters are summarized in Supplementary Section 6.2. Domain randomization is employed for all the learning stages. To acquire the exteroceptive information and the global location of the MAX robot in real-world experiments, we adopt a motion capture system and create an offline terrain map for the robots, and all real-world results reported in Figures~\ref{fig:pmc}-\ref{fig:sepmc} were obtained using the motion capture system. However, we have discussed and demonstrated in Section~\ref{sec:discussion} that replacing the motion capture system with onboard cameras is also achievable.  

\section*{DATA AVAILABILITY}
The raw motion data from the Labrador retriever together with the retargeted data for the MAX robot are available from Code Ocean at \url{https://doi.org/10.24433/CO.8441152.v3} and GitHub at \url{https://tencent-roboticsx.github.io/lifelike-agility-and-play/}. 
The raw motion clips are in .bvh format, and the retargeted data are organized in .txt files.

\section*{CODE AVAILABILITY}
\noindent
The codes are available in Code Ocean at \url{https://doi.org/10.24433/CO.8441152.v3} and GitHub at \url{https://tencent-roboticsx.github.io/lifelike-agility-and-play/}.

\section*{ACKNOWLEDGEMENT}
We would like to thank Shidi Li for his early contributions to motion retargeting. 
We would like to thank our colleagues in Tencent Robotics X and Tencent Cloud for providing constructive discussions and computing resources. We would like to thank the Labrador who wore the motion capture markers and performed movements for motion data collection.

\section*{AUTHOR CONTRIBUTIONS}
Lei Han organized the research project. Lei Han, Qingxu Zhu, Chong Zhang, Tingguang Li and He Zhang designed, implemented and experimented on various environmental settings, neural network architectures, algorithms, etc. Cheng Zhou, Tingguang Li and Chong Zhang collected the animal motion dataset. Lei Han and Yizheng Zhang iterated over multiple versions of the physics-based simulator and its settings.
Jiapeng Sheng, Yuzhen Liu, Yizheng Zhang, Tingguang Li, Qingxu Zhu and Lei Han completed the real robot experiments. 
Qingxu Zhu, Rui Zhao and Cheng Zhou contributed to improving the training infrastructure. 
Yuzhen Liu, Jie Li, Yufeng Zhang, Rui Wang, Wanchao Chi, Xiong Li, Yonghui Zhu, Lingzhu Xiang and Xiao Teng kept maintaining the robot hardware and software during the entire project. 
Lei Han wrote the paper with contributions from He Zhang, Chong Zhang, Qingxu Zhu, Tingguang Li and Jiapeng Sheng. Zhengyou Zhang provided general scope advice and consistently supported the team.

\bibliography{scibib}
\bibliographystyle{unsrt}

\clearpage

\section{Supplementary Material}

\subsection{Rewards and Training Details}
\label{sec:reward}

\subsubsection{Rewards and Training Details in Primitive-Level Training}

During training of the PMC, the policy aims to imitate the reference trajectory in motion clips. 
The tracking / imitation problem is formulated as a reinforcement learning problem with the tracking reward defined as below.
Similar to~\cite{RoboImitationPeng20}, the reward $r_t$ is defined as
\[
    r_t=w^{jp} r_t^{jp}+w^{jv} r_t^{jv}+w^{k} r_t^{k}+w^r r_t^r+w^v r_t^v,
\]
\[
    w^{jp}=0.6, w^{jv}=0.05, w^{k}=0.1, w^{r}=0.15, w^{v}=0.1,
\]
where the joint angle reward $r_t^{jp}$ and joint velocity reward $r_t^{jv}$ are defined as
\[
  r_t^{jp}=\exp \left[-1.0\left(\sum_j\left\|q_{t, j}^*-q_{t, j}\right\|^2\right)\right],
\]

\[
  r_t^{jv}=\exp \left[-0.1\left(\sum_j\left\|\dot{q}_{t, j}^*-\dot{q}_{t, j}\right\|^2\right)\right],
\]
and $q_{t, j}$ and $\dot{q}_{t, j}$ denote the joint angle and joint velocity of the j-th joint at time t; $q_{t, j}^*$ and $\dot{q}_{t, j}^*$ denote the target joint angle and joint velocity represented in motion clips.

The reward $r_t^{k}$ is defined to tracking the position of the key body parts in the robot. 
In our experiment, the key body parts are selected as four toes of the robot. 
Therefore, the $r_t^{k}$ is formulated as
\[
  r_t^{k}=\exp \left[-40.0\left(\sum_k\left\|p_{t, k}^*-p_{t, k}\right\|^2\right)\right],
\]
where $p_{t, k}$ denotes the position of the k-th toe relative to the root, whereas $p_{t,k}^*$ represents the position in motion clips.

The root position reward $r_t^r$ and root velocity reward $r_t^v$ are defined as

\[
  r_t^{r}=\exp \left(-20.0\left\|p_{t, r}^*-p_{t, r}\right\|^2 - 10.0 * \theta^2 \right),
\]
 
\[
  r_t^{v}=\exp \left(-2.0\left\|v_{t, r}^*-v_{t, r}\right\|^2 - 0.2 * \left\|\omega_{t, r}^*-\omega_{t, r}\right\|^2 \right),
\]
where $p_{t, r}$, $v_{t, r}$, and $\omega_{t, r}$ represent the position, linear velocity, and angular velocity of the root, respectively; $p_{t, r}^*$, $v_{t, r}^*$, and $\omega_{t, r}^*$ are the target values stored in motion clips; $\theta$ denotes the angle in the axis-angle representation of $R^* R^{-1}$, while $R^*$ represents the root orientation in motion clips and $R$ is the root orientation of the agent.

At the beginning of each episode, we randomly sample a motion clip from the dataset using prioritized sampling and set the motion clip as the reference motion. Then, we choose a uniformly random frame from this motion clip as the initial state and set the initial state of the legged robot according to this frame. The task in this episode is to track the remaining trajectories within this motion clip. Therefore, for each episode, the length varies according to the selected motion clip and the random initial frame. 
The episode will terminate once the robot reaches the end frame of the current motion clip. Additionally, to enhance data efficiency, early termination will also be activated if the robot falls or significantly deviates from the reference motion.

\begin{figure}[t]
\centering
\includegraphics[width=1.0\linewidth]{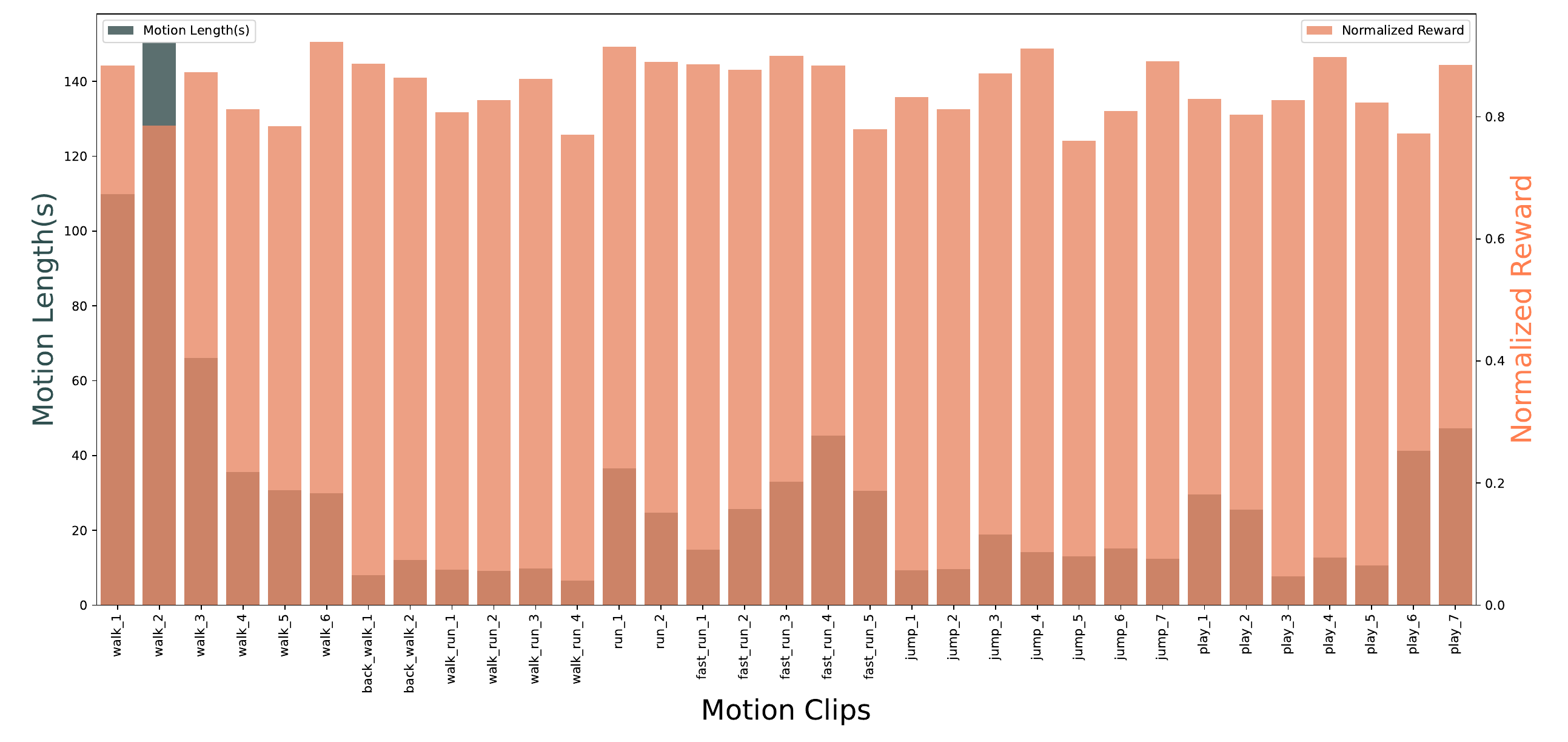}
{Figure S1: The distribution of motion clip length and normalized tracking reward.}
\end{figure}

The motion data contains 33 motion clips in total with variant lengths. Each motion clip is mirrored, and hence there are a total number of 66 motion clips.
In Figure~S1,
we plot the distribution of motion lengths (in seconds) and their associated tracking rewards (averaged over each clip and its mirrored version).
Generally, the tracking rewards reply on the difficulty of the reference motion clip. For example, motion clips that are with extraordinary agility like high jump might be harder to track. Thanks to the prioritized sampling scheme, we can achieve relative balanced performance over all the reference motion clips.

\subsubsection{Rewards and Training Details in Environmental-Level Training}
The EPMC is trained by controlling the policy to traverse diverse obstacles while maintaining given navigation direction and speed.
The navigation direction is determined by the target position in each scenario, while the speed is uniformly sampled from [0.5m/s, 3m/s] at the beginning of each episode for all tasks. The reward $r_t$ is defined as
\[
r_t = 0.5 * r_{\text {dir }}+ 0.5 * r_{\text {vel }},
\]
where $r_{dir}$ and $r_{vel}$ denote the navigation direction and speed following reward, respectively. The two terms are calculated as
\[
r_{dir}=\exp (-5|1-d \cdot \hat{d}|),
\]
\[
r_{vel}=\exp \left(-4\left|v^*-d \cdot v\right|\right),
\]
where $\hat{d}$ is the facing direction of the robot, and $d$ is the navigation direction command, while $v$ is the global linear velocity of the robot and $v^*$ is the navigation speed command.
For creeping, hurdle and block scenarios, $r_{dir}$ is calculated every time step, but
$r_{vel}$ is only counted for the average speed at the end of the episode as a sparse reward.
Our navigation reward is with compact structure consisting of only two terms, and it is the only task reward used in training the EPMCs for creeping, hurdles, and blocks.

For stairs, to discourage the robot from stepping on the edge of the stairs, we add an additional penalty when the toes are too close to the stair edge as
\[
r_{stair}=-\sum_{i=1}^4 I_i * 0.25,
\]
where $I_i$ is an indicator function that $I_i=1$ if the distance between the contact point of the $i$-th toe and the closest stair edge to the $i$-th toe is shorter than 5cm, otherwise $I_i=0$.

In Section~\ref{sec:discussion}, we mentioned that we have included an additional fall recovery environmental-level network. 
To incorporate the fall recovery skill, the objective is modified as
\[
    \min_{\bar{\pi}^{\text{EPMC}}}\ 
    E_{i\sim\{\text{flat,creeping, hurdle, block, stair,} \cdots\}}
    E_{\tau\sim\text{env}_i,\bar{\pi}^{\text{EPMC}}}
    \mathcal{KL}\left(\bar{\pi}^{\text{EPMC}} || \pi^{\text{teacher}_i} \right),
\]
where
\[
\pi^{\text{teacher}_i} = \mathbf{1}[s \notin \mathcal{R}] \pi^{\text{EPMC}_i} + \mathbf{1}[s \in \mathcal{R}] \pi^{recovery},
\]
and $\mathbf{1}[\cdot]$ is the indicator function, $\mathcal{R}$ denotes the set of states where the robot has fallen, and $\pi^{recovery}$ is the teacher recovery policy. The recovery policy is trained with random initial falling states. Specifically, to enrich the initial state of the robot, we reset the robot with random joint positions in the air and the robot falls freely for 1.0 second before taking action.
The reward $r_t$ in the fall recovery task is defined as
\[
r_t=w^or^o_t+w^{\omega}r^{\omega}_t+w^{jp}r^{jp}_t+w^{jv}r^{jv}_t,
\]
\[
w^o=80.0,w^{\omega}=-0.1,w^{jp}=-0.015,w^{jv}=-0.015,
\]
where the root orientation reward $r^o_t$ is defined as
\[
r^o_t=o^{eg}_{t,-1}-o^{eg}_{t-1,-1},
\]
$o^{eg}$ denote the unit gravity vector expressed in the base frame, and $-1$ indicates the last value of the vector;
the root angular velocity reward $r^{\omega}_t$ is defined as
\[
r^{\omega}_t=\left\|\omega_{t}\right\|^2
\]
where $\omega$ denotes the angular velocity of the robot body;
the joint angle $r^{jp}_t$ reward and joint velocity $r^{jq}_t$ reward are defined as
\[
r^{jp}_t=\sum_j\left\|p^N_{j}-p_{t, j}\right\|^2
\]
\[
r^{jq}_t=\sum_j\left\|q_{t, j}\right\|^2,
\]
where $p_{t, j}$ and $q_{t, j}$ denote the joint angle and joint velocity of the $j$-th joint at time $t$; $p^N_{j}$ denotes the excepted joint angles when the robot recovers from the fall.

For the entire training of multi-expert distillation, we used a single V100 GPU, and the student policy (uniform EPMC) only needs half an hour to achieve an average reward of 0.7, by noting that the average expert / teacher reward uniformly across all the EPMC tasks (including hurdles, creeping, blocks and stairs) is around 0.74. 
After 21 hours of training, the student policy obtains nearly the same average reward and perfect success rate for all the tasks as the teachers can achieve.
Figure~S2 shows the reward of the student policy during the training of multi-expert distillation.

\begin{figure}[t]
\centering
\includegraphics[width=1.0\linewidth]{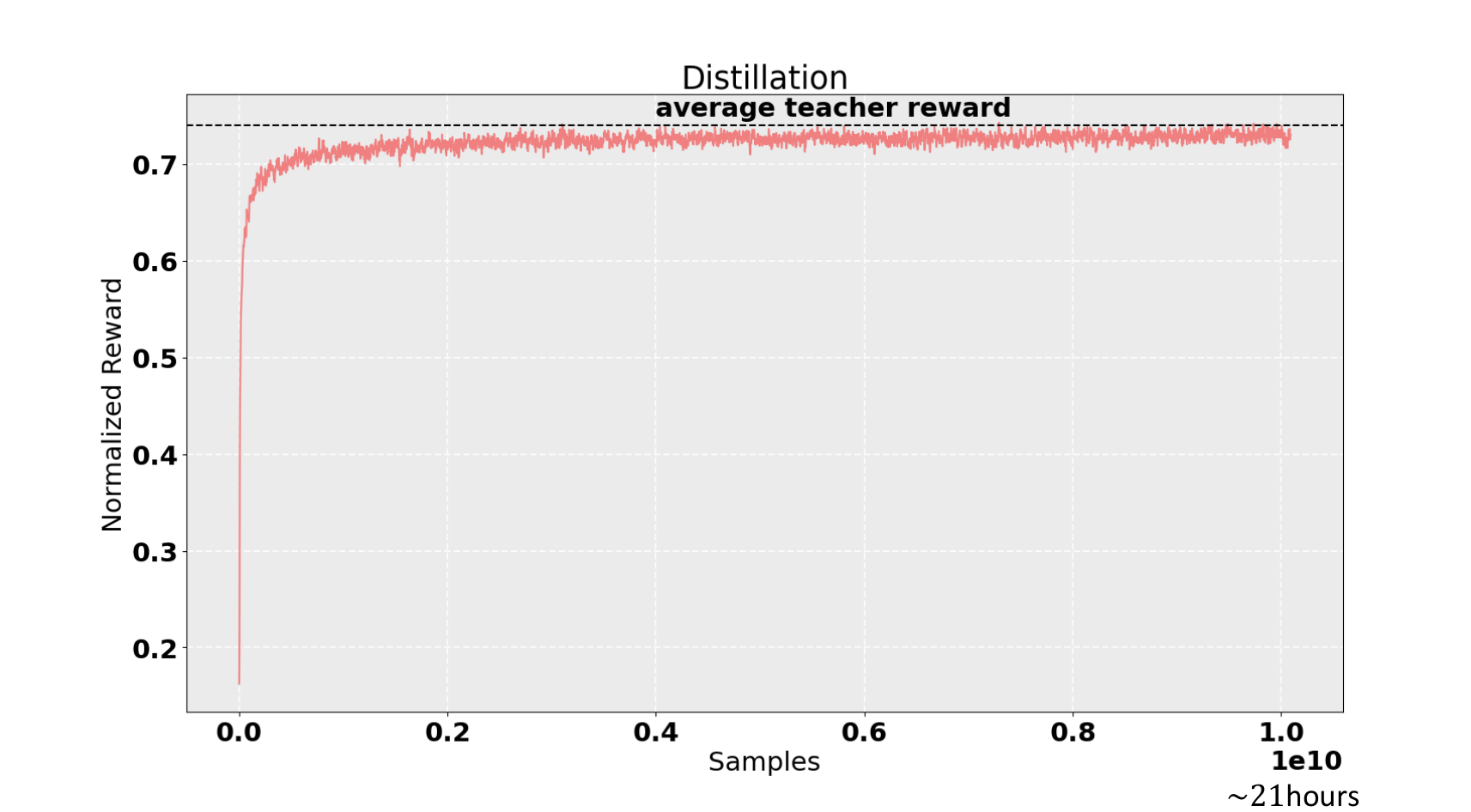}
{Figure S2: The average reward of the student policy (i.e., the uniform EPMC) during training of multi-expert distillation. We used a single NVIDIA TESLA V100 GPU.}
\end{figure}

\subsubsection{Rewards in Strategic-Level Training}
Sparse reward is used in the multi-agent Chase Tag Game. 
The robot receives a reward of $+1$ if it wins the game; otherwise, it receives $-1$. 
If the game is not terminated within 1000 time steps, both robots receive zero rewards. Figure~S3 shows the Elo scores during the PFSP training.

\subsection{Domain Randomization}
\label{sec:domain}
The lateral friction is uniformly randomized within $[0.4, 1.0]$. A random force is applied to the root of the robot with horizontal force uniformly sampled from $[0N, 50N]$ and vertical force sampled from $[0N, 10N]$ every 1.0s, and the forces last for 0.2s each time.
The torque limit is set to a random value in [15Nm, 18Nm] in simulation, and it is set to 15Nm in real-world deployment.
The invisible inflated cylinder is a trick used for training the EPMCs in simulation to encourage the agent to step far away from the edges to reduce the falling risk in reality.
Specifically, for all the training tasks containing obstacles, the edges of the obstacles are wrapped with invisible inflated cylinders with random radius in [0cm, 2cm], where `invisible' indicates that the wrapped cylinders are not observed by the robot while it is contactable.

Specific environmental randomization for each scenario in simulation is set as below.
For stairs, the heights and step widths are sampled from a distribution during training in simulation. The step heights are uniformly sampled from [0m, 0.18m] and the step widths are uniformly sampled from [0.34m, 0.4m]. 
For creeping tasks, the height of the narrow passage is uniformly sampled from [0.25m, 0.3m], the distance between any two consecutive passages is uniformly sampled from [1m, 3m], and the number of passages in one episode is uniformly sampled from [1, 10].
In real-world experiments, we set the height of the narrow passage as 0.25m.
For hurdles, the height of each hurdle is uniformly sampled from [0.05m, 0.15m], the distance between any two consecutive hurdles is uniformly sampled from [1m, 3m], and the number of hurdles in one episode is uniformly sampled from [1, 10].
In real-world experiments, we used hurdles with heights 0.1m and 0.15m.
For blocks, one set of blocks consists of a number of three to five blocks, each of which is with fixed width of 0.5m and height from \{0.1m, 0.25m, 0.4m\}. In each episode, the number of block sets is uniformly sampled from [1, 5].
In real-world experiments, we used four blocks with heights of 0.1m and 0.25m to compose the block set.
The configuration of each simulation environment can be viewed in Video~4.
For all scenarios, the distance between the two walls is randomized in [1m, 10m].

\subsection{Network Architecture and Hyperparameters}
\label{sec:net}
\subsubsection{Primitive-Level Training}
The action of the PMC is a vector of length 12, indicating the joint positions for the 12 actuators (each leg is controlled by 3 actuators).
The input $s_t^p$ of the PMC is a 135-dim vector stacking three frames of historical states, each of which concatenates the proprioception with last action.
The reference future trajectory $s_t^f$ is a 72-dim vector.
The value function, encoder and decoder are all parameterized as deep neural networks. 
The value function is modeled with a fully connected network with 2 hidden layers of $[256, 256]$ units. For the policy network, the encoder takes state $s_t$ and reference future motion $f_t$ as input and maps them to a latent variable $z_t^e$ using a fully connected network with 2 hidden layers of $[256, 256]$ units. 
The latent $z_t^e$ is quantized to $z_t^q$, and the decoder takes state $s_t$ and $z_t^q$ as input and maps them to a Gaussian distribution with a mean $\pi^{d}(a_t|s_t,z_t^q)$ and a trainable diagonal covariance matrix $\Sigma_\pi$. 
The decoder is also modeled with a fully connected network with 3 hidden layers of $[128, 256, 256]$ units. 
The hyperparameters used in this learning stage is summarized in Table~S1. 
GAE denotes the Generalized Advantage Estimate used in PPO~\cite{schulman2017proximal}.

\begin{table}[t]
\centering
\small
\caption*{Table S1: Hyperparameters for primitive-level training.}
\begin{tabular}{|l|l|} 
\hline
\textbf{Hyperparameter}                                & \textbf{Value}      \\ 
\hline
Number of codes $K$                                     & 256      \\ 
\hline
Code dimension $D$                                     & 32       \\ 
\hline
Commitment penalty coefficient $\beta$                             & 0.25       \\ 
\hline
GAE($\lambda$)                                         & 0.95     \\ 
\hline
Discount factor $\gamma$                               & 0.95     \\ 
\hline
Policy/value function learning rate                    & 0.00001  \\ 
\hline
Policy/value function batch size                   & 8192    \\ 
\hline
PPO clip threshold                                     & 0.2      \\ 
\hline
Prioritized sampling coefficient $\alpha_1$              & 3        \\
\hline
GPU used  &  2 NVIDIA TESLA V100 \\
\hline
Training time & 2 days \\
\hline
\end{tabular}
\end{table}

\begin{table}[t]
\centering
\small
\caption*{Table S2: Hyperparameters for environmental-level training.}
\begin{tabular}{|l|l|} 
\hline
\textbf{Hyperparameter}                                & \textbf{Value}      \\ 
\hline
GAIL reward weight for flat terrain                                 & 1.0     \\ 
\hline
GAE($\lambda$)                                         & 0.95     \\ 
\hline
Discount factor $\gamma$                               & 0.95     \\ 
\hline
Policy/value function learning rate                    & 0.00005  \\ 
\hline
Policy/value function batch size                   & 16384    \\ 
\hline
PPO clip threshold                                     & 0.1      \\ 
\hline
GPU used  & 4 NVIDIA TESLA V100 / scenario \\
\hline
Training time & 4 days / scenario \\
\hline
\end{tabular}
\end{table}

\subsubsection{Environmental-Level Training}
The primitive-level network reuses the pre-trained parameters of the decoder in VQ-PMC and keeps fixed in environmental-level training.
In addition to $s_t^p$, the environmental-level network additionally takes the exteroceptive observation $s_t^e$ as input, including a $25\times13$ 2D height map, a $25\times13$ 2D front depth map, and a 1D 128-dim vector from the ray sensor.
The two 2D maps are processed through four conv2d layers with sizes of 1x1@4, 4x4@4, 2x2@4 and 2x2@1, respectively. 
The output embeddings are then flatten and concatenated. 
The 1D 128-dim vector is processed by a cyclic conv1d layer with size 4@4, and three common conv1d layers with sizes of 4@4, 4@4 and 4@1, respectively. The output embedding is concatenated with the flatten 2D embeddings to compose the final embedding of $s_t^e$. 
The embeddings of $s_t^p$ and $s_t^e$ are then concatenated and input to a fully connected layer of size 256, connected by a LSTM layer with 32 hidden units. The output of the LSTM layer is then processed by a fully connected layer to output the index of the code. The hyperparameters used in this learning stage is depicted in Table~S2.

\subsubsection{Strategic-Level Training}
The environmental- and primitive-level networks reuse the pre-trained parameters and keep fixed in strategic-level training.
In addition to $s_t^p$ and $s_t^e$, the strategic-level network additionally takes the task information $s_t^{ta}$ as input, which is a vector concatenating a binary value indicating if the current agent is the chaser, a 7-dim vector indicating the information of the flag, and a 15-dim vector indicating the information of the opponent.
$s_t^{ta}$ is processed by 2 fully connected layers of size $[256, 256]$. 
The output embedding is concatenated with the embeddings of $s_t^p$ and $s_t^e$, and will be fed to a fully connected layer of size 256, connected by a LSTM layer with 32 hidden units. The output of the LSTM layer is then processed by a fully connected layer to output the navigation command. The hyperparameters used in the strategic-level training is provided in Table~S3.

\begin{table}[t]
\centering
\small
\caption*{Table S3: Hyperparameters for strategic-level training.}
\begin{tabular}{|l|l|} 
\hline
\textbf{Hyperparameter}                                & \textbf{Value}      \\ 
\hline
PFSP sampling coefficient $\alpha_2$              & 2.0        \\
\hline
GAE($\lambda$)                                         & 0.95     \\ 
\hline
Discount factor $\gamma$                               & 0.95     \\ 
\hline
Policy/value function learning rate                    & 0.00001  \\ 
\hline
Policy/value function batch size                   & 32768    \\ 
\hline
PPO clip threshold                                     & 0.1      \\ 
\hline
GPU used  & 4 NVIDIA TESLA V100  \\
\hline
Training time & 40 hours \\
\hline
\end{tabular}
\end{table}

\begin{figure}[t]
\centering
\includegraphics[width=0.5\linewidth]{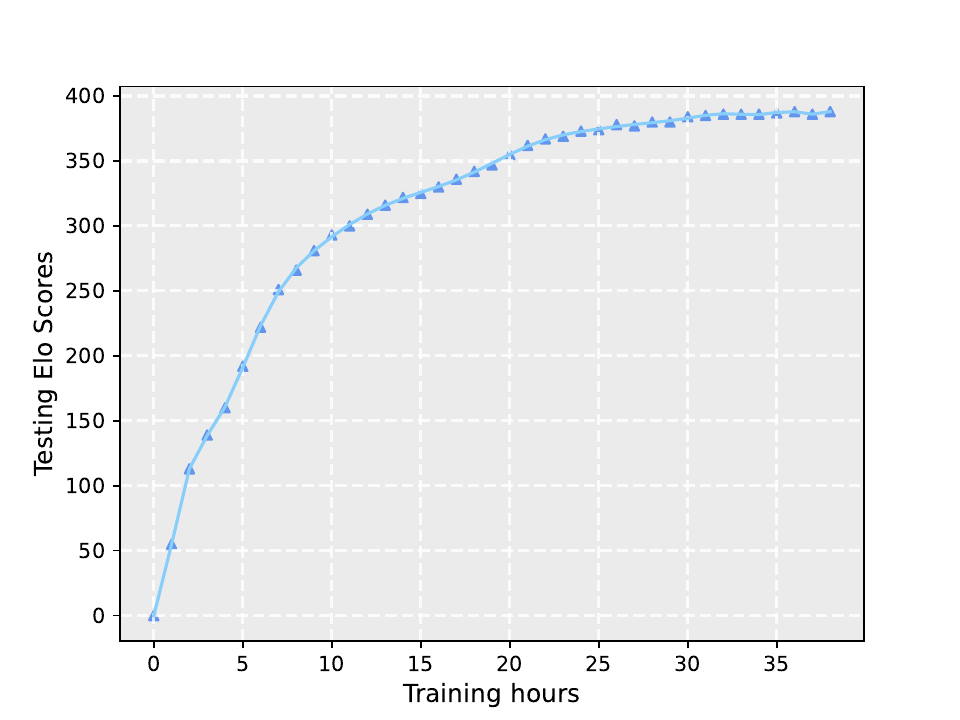}
{\\
Figure S3: The Elo scores for the stored models in the PFSP training progress.} 
\end{figure}

\subsection{Training Details of The Student Environment-Level Network}
\label{sec:student}
Similar to~\cite{doi:10.1126/scirobotics.abc5986,cheng2023extreme,zhuang2023robot}, we take an additional student-teacher distillation procedure, where our original trained controller is treated as the teacher, and we construct a student environmental-level network that takes only depth image from onboard camera as input, in addition to the proprioception and navigation command. The student environmental-level network also outputs a code to drive the primitive-level network, which exactly reuses the same frozen pre-trained PMC model. So, we only need to perform a supervised learning / distillation procedure to train the student environmental-level network given the teacher environmental-level network's output, i.e., the latent code, as labels. 
For the student environmental-level network, the raw depth image is of size $(240, 424)$, and it is down-sampled into a size of $(48, 64)$.
The down-sampled image is processed through four conv2d layers with sizes of 1x1@4, 4x4@4, 2x2@4 and 2x2@1, respectively.
The proprioception and navigation command are processed as the same as that of the teacher network. The image embedding is then flatten and concatenated with the other embeddings, and the concatenation is processed by the same successive layers as those of the teacher network.
For the real MAX robot, we install an Intel RealSense Depth Camera D455 at the head of the robot. 
To reduce the gap between the depth image from the Intel RealSense Depth Camera and the image obtained in simulation, we follow~\cite{zhuang2023robot} to apply Gaussian noise and patch noise to the image in simulation during the distillation training.
Figure~S4 shows the comparison between the images obtained from the real Intel RealSense Depth Camera and in simulation.
\begin{figure}[t]
\centering
\includegraphics[width=0.6\linewidth]{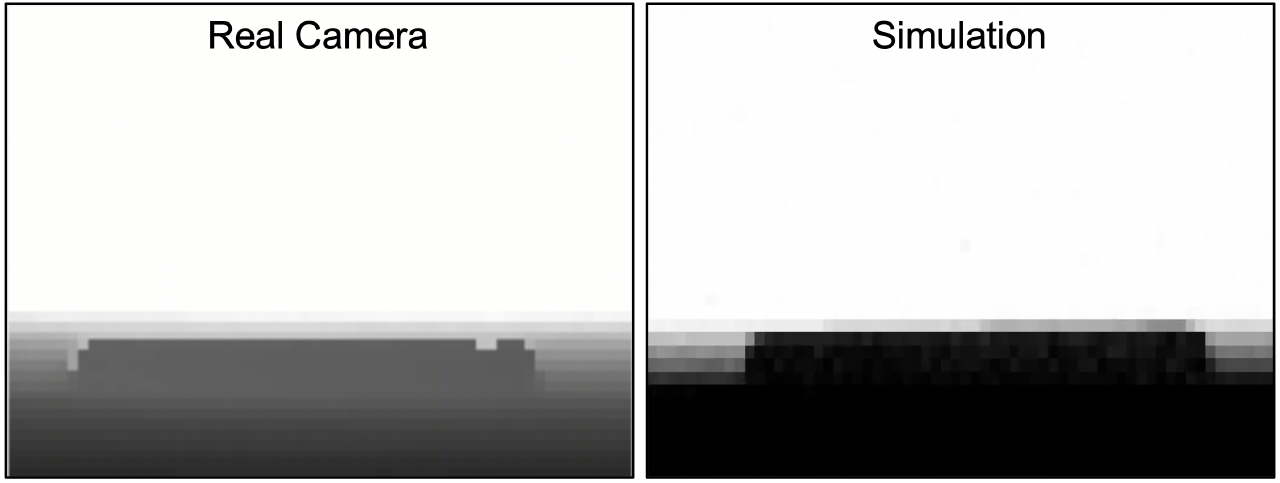}
{\\
\small
Figure S4: Comparison between the images obtained from the real camera and in simulation, when there is a hurdle with 10cm height in front of the robot.
}
\end{figure}
We can observe that these depth images are very similar, and this facilitates the sim2real transfer.  

Moreover, the original controller takes the root linear velocity as input to complete the tasks, where the root linear velocity is obtained by the motion capture system. For the student network, we still include the root linear velocity in the input, while we estimate it by training a small additional neural network instead of using motion capture. 
The small network takes the historical joint positions, joint velocities, root angular velocity and gravity vector as input, and it estimates the root linear velocity using a supervised loss given the true value in simulation. The root linear velocity estimation loss plus the distillation loss composes the final training loss.
The small neural network is a fully connected layer of size 256, connected by a LSTM layer with 32 hidden units. The output of the LSTM layer is then processed by a fully connected layer to output the estimation of the root linear velocity.
The training can achieve convergence within 3 hours using a single TESLA V100 GPU.

We deploy the trained student environmental-level network in the MAX robot equipped with depth camera in a zero-shot manner.
Video~6 shows the performance in the hurdle scenarios, where we deploy the MAX robot in a dark room and at an open square, both locations without motion capture systems.
We can observe from the video that the MAX robot with onboard camera reproduces the lifelike agility as the initial policy deployed using motion capture.

\subsection{Comparison with Concurrent Methods}
\label{sec:compare}
We first provide a summary for the recent concurrent methods.
\begin{itemize}
    \item Barkour~\cite{caluwaerts2023barkour}: the main motivation of Barkour is to set up a benchmark race competition with both fixed physical obstacle configuration and its digital version in simulation. To solve this race competition, three specialist policies are trained at the first stage to solve fast omni-directional walking on uneven terrain, climbing up and down a slope, and jumping over a board, separately. 
    Then, a higher-level navigation controller is trained to select an appropriate specialist policy based on the location of the robot in the fixed environment. At the second stage, the specialist policies are distilled into a single uniform transformer-based locomotion policy. The exteroception is modeled as a heightfield in the vicinity of the robot in both simulation and reality, where the perception is acquired via the motion capture system in reality, similar to the exteroception used in our method. Careful reward engineering (including 14 terms and handcraft waypoints along the race trajectory), terrain curriculum design and domain randomization are conducted to achieve the final performance. The robot hardware is developed in-house and is similar in size to the Unitree A1 and MIT Mini-Cheetah.
    \item CAJun~\cite{yang2023cajun}: CAJun considers a specific task of continuous jumping. It utilizes a hierarchical framework that consists of a low-level controller using gait generator and QP-based optimization to output motor commands, and a high-level neural network controller outputting the stepping frequency, the swing foot residual, and the desired base velocity. The perception focuses on the proprioceptive information and task information including the jump phase and the target landing positions. Exteroceptive information is not considered. Reward function is carefully designed for the specific task with 9 terms. 
    One benefit is that with an optimal control based low-level module, the method is efficient to solve the task with 20 minutes of training the high-level controller. The robot hardware used in the experiments is the Unitree Go1.
    \item ANYmal Parkour~\cite{hoeller2023anymal}: the pipeline of ANYmal Parkour includes three interconnected components that are a perception module, a locomotion module and a navigation module. The perception module receives point cloud input from the onboard cameras and the LiDAR and estimates the terrain height around the robot. Five specialist policies are trained to tackle different terrain and obstacles, and a higher-level navigation policy chooses which specialist policy to use. A total number of eight neural networks are separately tuning. Careful reward engineering (the locomotion rewards for training specialist policies contain 16 terms), terrain curriculum and domain randomization are applied. The robot hardware used is the ANYmal D developed by ETH Zurich.
    \item Parkour Learning~\cite{zhuang2023robot}: the core innovation is a specific two-stage curriculum pipeline for traversing obstacles. At the first stage, the contact between the robot and the obstacles is deactivated, while the volume of soft penetration of the body into the obstacles are computed and used as a negative reward to penalize the training. At the second stage, the contact is activated and the policy is initialized by the model from the first stage to facilitate the training. Privileged information and domain adaption are used in training. Five specialist policies are trained separately to handle climbing, leaping, crawling, and tilting scenarios. For each scenario, a specific reward configuration is designed with 6 terms, resulting in a total number of 24 specific reward hyperparameters. Unitree A1 and Unitree Go1 with onboard visual sensors are deployed in real-world experiments.
    \item Extreme Parkour~\cite{cheng2023extreme}: the extreme parkour method uses a teacher-student distillation framework that is similar to~\cite{doi:10.1126/scirobotics.abc5986}. The teacher policy receives scandots (treated as privileged information) around the robot to perfectly describe the terrain in simulation, while the student takes depth image from onboard sensors equipped on the robot as input to learn the teacher's action. Despite this, a main innovation claimed in the study is a unified reward designed for parkour setting. That is, there are a sequence of predefined waypoints to navigate the robot and a compact reward formulation (with only 3 items) is defined according to each of these waypoint conditioning on the current robot position. Terrain curriculum is adopted to accelerate the learning process. Unitree A1 with onboard sensors is used in real-world experiments.
\end{itemize}

For simplicity, we use Barkour, CAJun, ANYmal, Parkour and Extreme to represent the five methods, respectively. Overall, Barkour, ANYmal, Parkour, Extreme and ours consider end-to-end neural networks, and these methods aim to learn a number of challenging traversing skills that are not common in previous studies. These skills are compressed into either a uniform policy or a mixture-of-expert (MoE) policy with a navigation module. CAJun focuses on the task of continuous jumping and proposes a hybrid control method by combing neural network and optimal control. 
The major differences between these methods and ours are that
\begin{itemize}
    \item we embrace the concept of pretraining and organizing the knowledge into hierarchies according to perception and control levels. Each level of knowledge can be preserved, reused and enriched. Note that Barkour, ANYmal, Parkour and Extreme learn each skill separately and then compress the skills into a uniform policy or employ MoE. The learning progresses of the skills are isolated and do not benefit from each other. In fact, for example, jumping over gaps and jumping over hurdles may share common movement elements, and learning one skill should benefit the learning of others. Abstracting a primitive-level network can help shape common primitive knowledge, and therefore we defer the learning of skills to the environmental-level control by reusing a common primitive-level layer;
    \item we propose to pretrain primitive behaviors from animal motions using a generative model. The generated movements are naturally lifelike, and we do not need to focus on carefully hand-designed rewards for naturalistic and energy-efficient consideration, by noting that a number of complex reward items focusing on balancing such considerations are involved in all the other methods while our reward function for skill learning (EPMC) only contains two terms (please refer to Table~S4).
    Meanwhile, we advocate the use of sparse rewards, the reasons for which we have explained in Section 5.2 of the main paper. 
    Sparse rewards can provide unbiased description of the tasks, simplify the reward structure and reduce the labor of dense reward design, while it makes the task much more challenging. 
    Reusing a pretrained primitive-level layer from animal data significantly facilitates the learning of the environmental-level network to handle sparse rewards problem.
    Furthermore, it is worth mentioning that we did not perform any curriculum learning procedures, because by reusing a powerful pretrained PMC, it is efficient to solve various considered tasks in a straightforward manner. On the contrast, most of these recent methods rely on carefully designed curriculum learning scheme;
    \item our approach is the only work considering and experimenting with multi-agent games. All the related approaches focus on locomotion capability, while in addition to that we further investigate the learning of strategic-level knowledge. This also benefits from the proposed hierarchical framework, which allows us to focus on the strategic-level training conditioning on the robust and reusable environmental- and primitive-level networks. Moreover, we designed a novel multi-agent Chase Tag Game and demonstrated interesting strategic-level policies via self-play;
    \item contribution to the field: the source codes, pretrained model parameters, and all animal motion data will be released to the community that future researches can conveniently reuse the pretrained models and motion data without reproducing our training pipeline.
\end{itemize}
Table~S4 provides a comprehensive analysis and comparison among all the methods, from which our approach keeps novel compared with these state-of-the-art methods by considering the above points.

\begin{landscape}
\begin{table}[t]
\centering
\footnotesize
\caption*{Table S4: A comprehensive summary and comparison of recent concurrent methods.}
\scalebox{0.8}{
\begin{tabular}{l|c|c|c|c|c||c} 
\hline
 & Barkour & CAJun & ANYmal & Parkour & Extreme & \textbf{Ours} \\ 
\hline
\hline
\multirow{3}{*}{Policy} & \multirow{3}{*}{A uniform policy} & \multirow{3}{*}{A single policy} & Multiple specialist & \multirow{2}{*}{Multiple specialist} & \multirow{3}{*}{A uniform policy} & \multirow{3}{*}{A uniform policy} \\ 
 & & & policies and a & \multirow{2}{*}{policies} & & \\ 
 & & & navigation policy & & & \\ 
\hline
\multirow{3}{*}{Framework} & \multirow{2}{*}{Multi-expert} & Hybrid method & Multi-module: & \multirow{2}{*}{Two-stage} & \multirow{2}{*}{Teacher-student} & Hierarchical: \\
& \multirow{2}{*}{distillation} & with an NN layer & perception, locomotion & \multirow{2}{*}{curriculum} & \multirow{2}{*}{distillation} & primitive, environmental \\
& & and a QP-based layer & and navigation & & & and strategic levels \\
\hline
\multirow{3}{*}{Task} & \multirow{2}{*}{Single agent} & \multirow{2}{*}{Single agent} & \multirow{2}{*}{Single agent} & \multirow{2}{*}{Single agent} & \multirow{2}{*}{Single agent} & Navigation and \\
 & \multirow{2}{*}{navigation} & \multirow{2}{*}{jump} & \multirow{2}{*}{navigation} & \multirow{2}{*}{navigation} & \multirow{2}{*}{navigation} & a multi-agent \\
 & & & & & & Chase Tag Game \\
\hline
\multirow{6}{*}{Skills} & Climb slope & Continuous jump & Jump over gaps & Jump over gaps & Jump over gaps & Jump over gaps \\ 
 & Jump over a board & & Climb high obstacles & Climb obstacles & Jump over hurdles & Jump over hurdles \\ 
 & & & Crawl & Crawl & Run over tilted ramp & Crawl \\ 
 & & & & Tilt & Climb on wide steps & Climb dense stairs \\ 
 & & & & & & Jump over consecutive \\ 
 & & & & & & high blocks \\ 
\hline
Animal Data & \ding{55} & \ding{55} & \ding{55} & \ding{55} & \ding{55} & \checkmark \\ 
\hline
Lifelike & \ding{55} & \ding{55} & \ding{55} & \ding{55} & \ding{55} & \checkmark \\ 
\hline
\multirow{2}{*}{Skill Reward} & 14 terms and  & \multirow{2}{*}{9 terms} & \multirow{2}{*}{16 terms} & \multirow{2}{*}{6 terms} & 3 terms and & \multirow{2}{*}{2 terms} \\
 & predefined waypoints  & & & & predefined waypoints & \\
\hline
Hand-designed & \multirow{2}{*}{\checkmark} & \multirow{2}{*}{\ding{55}} & \multirow{2}{*}{\checkmark} & \multirow{2}{*}{\checkmark} & \multirow{2}{*}{\checkmark} & \multirow{2}{*}{\ding{55}} \\
Curriculum & & & & & & \\ 
\hline
\multirow{2}{*}{Hardware} & Developed & \multirow{2}{*}{Unitree Go1} & \multirow{2}{*}{ANYmal D} & \multirow{2}{*}{Unitree A1 and Go1} & \multirow{2}{*}{Unitree A1} & Developed \\ 
 & in-house & & & & & in-house \\ 
\hline
Exteroceptive & Terrain map via & \multirow{2}{*}{NA} & Terrain map via & Depth image via & Depth image via & MoCap system \\
Sensors & MoCap system & & onboard sensors & onboard sensors & onboard sensors & or depth camera \\
\hline
Weight & 11.5 kg & 12 kg & 55 kg & 12 kg & 12 kg & 14 kg \\ 
\hline
\multirow{2}{*}{Peak Torque} & \multirow{2}{*}{12 Nm} & \multirow{2}{*}{23.7$\sim$35.55 Nm} & \multirow{2}{*}{85 Nm} & 33.5 Nm and & \multirow{2}{*}{33.5 Nm} & \multirow{2}{*}{30 Nm} \\ 
 & & & & 23.7$\sim$35.55 Nm & & \\ 
\hline
\end{tabular}
}
\end{table}
\end{landscape}

\end{document}